\documentclass[letterpaper, 10 pt, conference]{ieeeconf}  

\IEEEoverridecommandlockouts                              

\usepackage{graphics} 
\usepackage{epsfig} 
\usepackage{times} 
\usepackage{amsmath} 
\usepackage{amssymb}  
\usepackage{newtxmath} 
\usepackage{graphicx}
\usepackage{cuted}
\usepackage{caption} 
\usepackage{svg}
\usepackage{url}
\usepackage{comment}
\usepackage{booktabs}
\usepackage{array}
\usepackage{tikz}
\usetikzlibrary{calc}
\usepackage{calc}
\usepackage{arydshln}
\usepackage[export]{adjustbox}

\newcommand{\errorellipse}[6]{%
\begin{tikzpicture}[baseline=(img.base)]
    \node[inner sep=0] (img) {\includegraphics[width=0.2\linewidth]{#1}};
    \begin{scope}[
        shift={(img.south west)},
        x={($(img.south east)-(img.south west)$)},
        y={($(img.north west)-(img.south west)$)}
    ]
        \draw[red, dotted, line width=1.2pt, rotate around={#6:(#2,#3)}]
        (#2,#3) ellipse (#4 and #5);
    \end{scope}
\end{tikzpicture}%
}
\newcommand{\errorellipsevar}[7]{%
\begin{tikzpicture}[baseline=(img.base)]
    \node[inner sep=0] (img)
    {\includegraphics[width=#7\linewidth]{#1}};
    \begin{scope}[
        shift={(img.south west)},
        x={($(img.south east)-(img.south west)$)},
        y={($(img.north west)-(img.south west)$)}
    ]
        \draw[
            red,
            dotted,
            line width=1.2pt,
            rotate around={#6:(#2,#3)}
        ]
        (#2,#3) ellipse (#4 and #5);
    \end{scope}
\end{tikzpicture}%
}

\newcommand{\rowlabel}[2]{%
\parbox[b][\heightof{\includegraphics[width=0.2\linewidth]{#2}}][c]{1.2em}{%
\centering\rotatebox{90}{\textbf{#1}}%
}%
}

\title{\LARGE \bf
InfiNoVA: \ensuremath{\infty} Novel View Augmentation for Viewpoint Invariant Robot Policies
}

\author{Sai Puneeth Reddy Gottam, Elmar Rueckert$^{\dagger}$ and Vedant Dave$^{\dagger}$
\thanks{$^{\dagger}$Equal Advising.}%
\thanks{All authors are associated with the Chair of Cyber Physical Systems,}%
\thanks{Montanuniversit\"at Leoben, Austria.}%
\thanks{Corresponding author: sai.gottam@unileoben.ac.at}%
}

\usepackage{fancyhdr}

\begin{document}

\bstctlcite{IEEEexample:BSTcontrol}

\maketitle
\thispagestyle{empty}
\pagestyle{empty}

\begin{strip}
\centering
\includegraphics[width=0.95\textwidth]{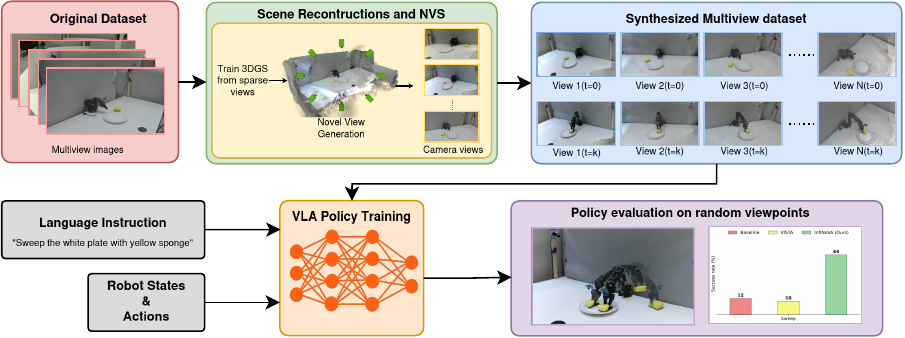}
\captionof{figure}{Our framework (InfiNoVA) creates 3D scene from Gaussian splatting and renders novel views for training VLA policies.}
\label{fig:main}
\vspace{-0.5cm}
\end{strip}



\begin{abstract}
Vision-Language-Action (VLA) policies often rely strongly on the camera viewpoints seen during training, causing substantial performance degradation when deployed from unseen perspectives. Collecting demonstrations from sufficiently diverse physical viewpoints is expensive and still provides only sparse coverage of the viewpoint space. We introduce InfiNoVA, a data-augmentation framework that converts synchronized multi-camera demonstrations into a dense distribution of geometrically consistent training views. InfiNoVA reconstructs each manipulation trajectory as a time-varying 3D Gaussian representation and renders novel observations from sampled camera poses while preserving the original state--action correspondence. This explicit scene representation improves frame-level fidelity and temporal consistency while reducing task-critical hallucinations observed in generative novel-view synthesis. Across four real-world manipulation tasks, policies trained with InfiNoVA achieve $5.4\times$ higher average success under unseen randomized viewpoints than both VISTA-based augmentation and the unaugmented policy. InfiNoVA further achieves $1.7\times$ higher success than training directly on all five physical camera views. These results show that
dense, geometrically grounded viewpoint augmentation provides a practical route toward camera-robust robot policies without modifying the underlying policy architecture. Project page: \url{https://infi-nova.github.io}.
\end{abstract}

\section{Introduction}

Vision-Language-Action models (VLAs) have rapidly emerged as a promising interface between large-scale visual-language pretraining and robotic control. By conditioning action prediction on visual observations and language instructions, recent models such as RT-2~\cite{zitkovich2023rt}, OpenVLA~\cite{kimopenvla}, $\pi_0$~\cite{black2024pi_0}, and SmolVLA~\cite{shukor2025smolvla} demonstrate that pretrained multimodal representations can be adapted to robotic manipulation, enabling policies that interpret task intent, perceive objects, and produce actions from raw sensory input.

However, learning such robotic manipulation policies requires large and diverse datasets. This data must capture not only visual observations and language instructions, but also a wide range of object interactions, camera viewpoints, and environmental variations. Collecting such diverse demonstrations in the real world is costly, time-consuming, and difficult to scale. Such data collection often requires human teleoperation, multiple expert demonstrations, repeated task execution, and extensive effort to cover multiple environmental configurations. As a result, most datasets cover only a limited subset of possible viewpoints, lighting conditions, backgrounds, object configurations, and distractor patterns. While simulated environments offer a scalable alternative for collecting enormous amount of data, policies trained in simulation struggle to transfer directly to real hardware due to the sim-to-real gap, caused by discrepancies in visual fidelity, contact dynamics, and unmodeled physical variations~\cite{peng2018sim}. Consequently, policies remain brittle when deployed under real-world conditions that differ from their training setups, particularly because viewpoint and appearance changes can substantially alter the visual evidence available to a VLA without changing the task itself. Recent evaluations~\cite{fei2025libero,wang2025vlatest,gao2026taxonomy, 10611331, pumacay2024the,li2607camvla} show that current VLAs degrade substantially under shifts in camera pose suggesting that they overly rely on camera-specific visual correlations rather than the task structure itself. In particular, LIBERO-Plus~\cite{fei2025libero} reports an average relative performance drop of approximately 69.6\% under small camera-viewpoint perturbations across multiple VLA policies, highlighting camera shift as a major source of deployment failure.

A more direct way to scale viewpoint diversity is to physically add cameras. Multi-camera rigs capture the same demonstration from several fixed viewpoints simultaneously, giving policies genuine exposure to different camera poses during training. This improves robustness over single-view setups~\cite{pmlr-v305-yang25b}, but scaling further quickly becomes expensive as each additional viewpoint requires extra hardware, calibration, and synchronization, and even a well-instrumented rig only samples a finite, fixed set of positions. A policy trained on such multi-camera data can still overfit to those specific projections, and may fail when the deployment camera lies between, outside, or slightly shifted from the training views, a limitation we further examine in our multi-camera study (Sec.~\ref{subsec:multicam}).

To bypass physical hardware limits, recent work synthesizes new views. Novel-view synthesis (NVS) augmentation~\cite{tian2025view,zhou2023nerf,heo2026anycamvla} removes the hardware bottleneck entirely, generating additional viewpoints directly from existing demonstrations. However, approaches that synthesize views from a single source observation become increasingly challenging as the target viewpoint moves away from the input view, where previously occluded geometry, unseen object surfaces, and background regions must be inferred. Explicit 3D reconstruction methods~\cite{YangS-RSS-25} avoid this by building a full 3D scene representation, offering stronger geometric control, but rely on additional scene-editing and alignment stages to insert or manipulate objects within the reconstruction

\begin{figure}
    \centering
    \includegraphics[width=\linewidth]{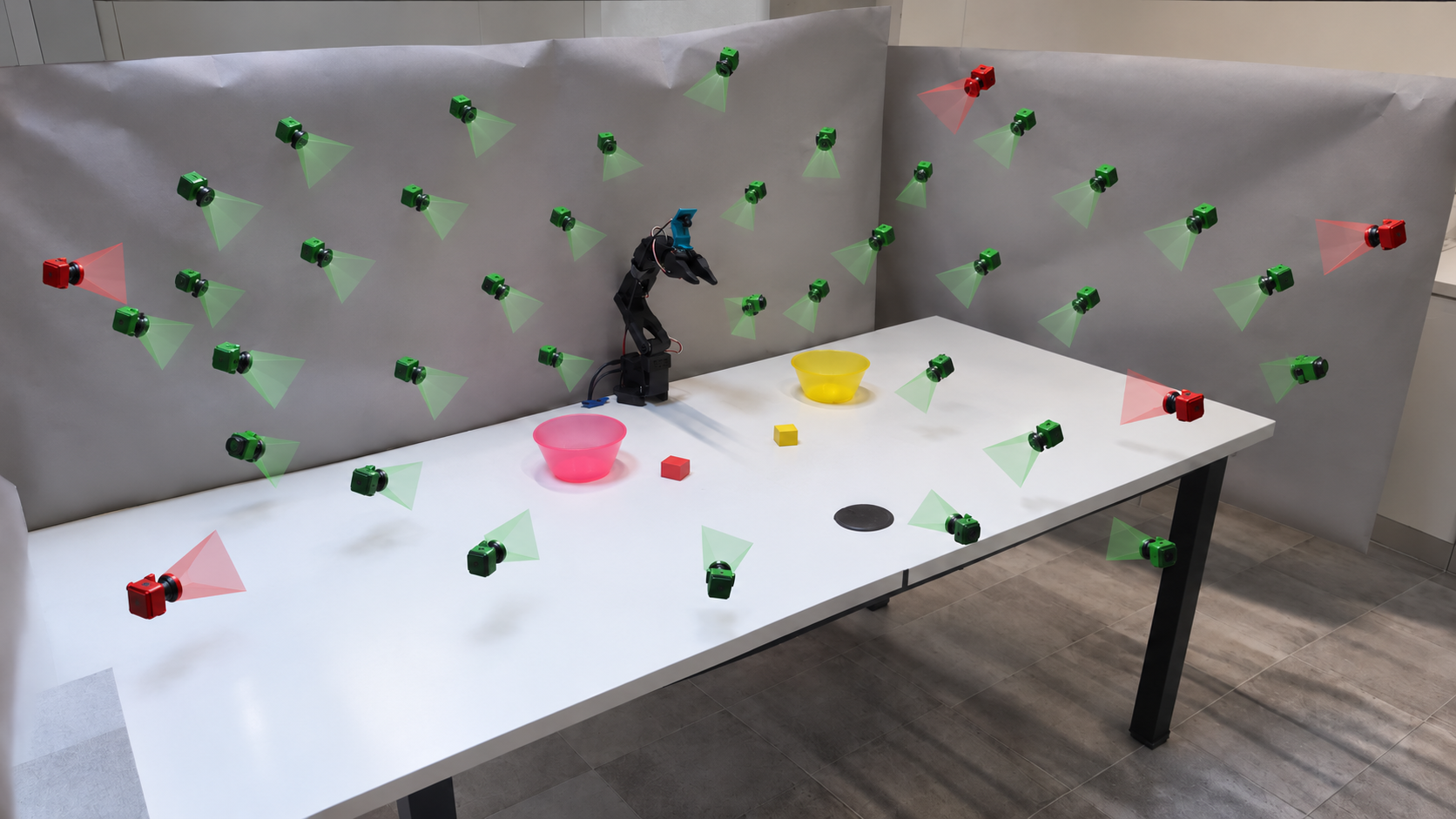}
    \caption{Multi-view reconstruction and novel-view sampling. Red cameras are used for reconstruction; green cameras denote sampled novel views.}
    \label{fig:idea}
\end{figure}

 The central premise of our work is that a VLA should not learn actions from a small set of fixed, camera-conditioned inputs, but from a family of geometrically valid observations that preserve the same underlying state-action relation. Decoupling action learning from static camera perspectives enables the policy to generalize seamlessly across varying viewpoints. We therefore treat synchronized multi-camera observations not merely as additional training inputs, but as the basis for constructing a continuous viewpoint augmentation space. Given a manipulation trajectory recorded from five cameras, we reconstruct a time-varying 3D Gaussian representation using InstantSplat++~\cite{fan2025instantsplatsparseviewgaussiansplatting}, whose pose-free formulation avoids requiring pre-calibrated camera extrinsics for scene reconstruction. The resulting renderable sequence remains temporally aligned with the original trajectory and action labels, allowing additional observations to be generated without relabeling the demonstration. Because the reconstruction is grounded in multiple synchronized real views, we can sample a dense set of virtual camera poses varying in azimuth, elevation, and radial distance within the observed workspace (Fig.~\ref{fig:idea}). This transforms a small set of physical camera observations into a substantially broader distribution of geometrically consistent training viewpoints, without requiring additional cameras or relying solely on single-view extrapolation.

\section{Related Work}

\subsection{Spatial Grounding in Vision-Language-Action Models}
Vision-language-action models (VLAs) extend vision-language pretraining to robotic control, mapping visual observations and language instructions to action sequences and demonstrating strong semantic generalization and cross-task transfer~\cite{zitkovich2023rt,kimopenvla,black2024pi_0,shukor2025smolvla,o2024open,team2024octo}. However, recent evaluations reveal substantial performance degradation under visual and camera-pose perturbations, highlighting limited spatial grounding across viewpoints~\cite{10611331,pumacay2024the}. Recent methods address this limitation by incorporating geometric information through 3D representations, point clouds, pretrained geometric features, or explicit camera extrinsics~\cite{qu2025spatialvla,sun2025geovla,abouzeid2025geoaware,jiang2026knowyourcamera, zhang2026grounding}. While these approaches improve viewpoint generalization, they often require architectural modifications, additional geometric inputs, or known/estimated camera calibration. CamVLA~\cite{li2607camvla} instead learns a camera-to-base transformation to avoid deployment-time calibration, but remains limited to a single third-person camera and struggles under large viewpoint changes. In contrast, our work addresses viewpoint robustness at the data level by augmenting existing demonstrations with geometrically consistent novel views, without modifying the policy architecture or requiring camera calibration at deployment.

\subsection{Generative Data Augmentation for Robot Generalization}
Generative augmentation has been widely used to improve robot-policy robustness without collecting additional demonstrations. Early domain-randomization methods varied colors, textures, and simulator appearance~\cite{tobin2017domain,alghonaim2021benchmarking}, while recent approaches use image-text models, diffusion, segmentation, and scene-level generation to diversify objects, backgrounds, distractors, and scene context~\cite{Chen-RSS-23,Yu-RSS-23,chen2025semantically,teoh2024green,yuan2025roboengine,wang2026roboaug}. These methods primarily broaden the appearance and scene distribution while keeping the underlying camera projection fixed. In contrast, our approach expands viewpoint diversity by rendering the same underlying state-action trajectory from novel camera poses while retaining alignment with the original action labels.

\subsection{Novel-View Synthesis for Robust Robotic Policies}
Novel-view synthesis (NVS) provides a direct mechanism for augmenting robot demonstrations with unseen viewpoints. VISTA~\cite{tian2025view} uses single-image diffusion-based NVS to synthesize alternate training views, but its viewpoint range is limited by hallucination of disoccluded geometry and unseen scene content. AnyCamVLA~\cite{heo2026anycamvla} instead performs NVS at test time to transform deployment observations into the policy's training viewpoint, but relies on camera calibration or pose estimation and on accurate real-time synthesis. Yang et. al.~\cite{YangS-RSS-25} uses Gaussian splatting for viewpoint and appearance augmentation, but requires scene decomposition and geometric alignment for comprehensive object and embodiment editing. Unlike these existing methods, we instead perform training-time viewpoint expansion from real multi-camera observations, reconstructing a scene-specific 3D representation with InstantSplat++. This provides geometrically grounded novel views across a wider range of poses while avoiding single-image hallucination, test-time calibration requirements, and the scene-decomposition overhead of comprehensive scene-editing pipelines.

\section{Approach}

\subsection{Problem Formulation}

We consider robotic manipulation policies conditioned on visual observations and language instructions. A demonstration trajectory is represented as
\(\tau={(o_t,l,a_t)}_{t=1}^{T}\), where \(o_t\) is the visual observation, \(l\) the language instruction, and \(a_t\) the expert action at timestep \(t\). Our goal is to learn a Vision-Language-Action policy \(\pi(a_t\mid o_t,l)\) that generalizes across camera viewpoints.

Training demonstrations typically cover only a small set of camera poses, causing policies to become sensitive to viewpoint-specific visual correlations. We therefore augment an original demonstration dataset \(\mathbb{D}\) with novel-view observations to obtain \(\mathbb{D}_{nv}\). For each sample \((o_t,l,a_t)\), we generate an observation \((\tilde{o}_t)\) from a different camera pose while preserving the underlying scene state and action semantics, such that \((\tilde{o}_t,l,a_t)\) remains a valid training sample.

We generate these observations from a reconstructed 3D Gaussian representation of each demonstration, enabling geometrically consistent novel-view rendering without collecting additional robot trajectories. The resulting dataset exposes the policy to a broader camera-view distribution and encourages generalization to viewpoints unseen during training.




\subsection{3D scene reconstruction and Novel View Synthesis}

As shown in the Fig~\ref{fig:idea}, our framework reconstructs the tabletop manipulation scene from synchronized multi-view observations. 
We use a five-camera setup to capture the workspace from different viewpoints, providing complementary visual evidence of the robot arm, gripper, tabletop, and manipulated objects. For each timestep \(t\), we collect the synchronized set of observations $\mathcal{O}_t = \{o_t^1, o_t^2, o_t^3, o_t^4, o_t^5\}$, where \(o_t^i\) denotes the image captured by the \(i\)-th camera at timestep \(t\). The multi-view image set \(\mathcal{O}_t\) is given as input to Instantsplat++~\cite{fan2025instantsplatsparseviewgaussiansplatting}, which reconstructs a Gaussian Splatting representation of the scene $ G_t = f_{\text{GS}}(\mathcal{O}_t).$

We use Instantsplat++ because it provides fast feed-forward sparse-view reconstruction and does not require externally provided camera parameters. 
This pose-free property is useful for robotic demonstration data, where accurate camera calibration may not always be available. 
The reconstructed representation \(G_t\) captures the task-relevant scene, including the tabletop workspace, robot arm, gripper, and manipulated objects.
After reconstructing the scene at timestep \(t\), we render novel views from a set of virtual camera poses \(\mathcal{C} = \{c_1, c_2, \dots, c_K\}\) placed around the workspace $\tilde{o}_t^k = R(G_t, c_k)$, where \(R\) denotes the rendering function and \(\tilde{o}_t^k\) is the rendered observation from virtual camera pose \(c_k\). Repeating this process for all timesteps produces novel-view trajectories that preserve the original robot motion and object configuration. Therefore, each rendered observation can be paired with the original language instruction and action label.

The multi-view setting is important for reducing hallucinations in novel-view synthesis. Single-view approaches are highly under constrained because they must infer unseen geometry, depth, and occluded regions from only one image. 
This can lead to unrealistic or task-inconsistent outputs, particularly around the gripper, robot arm, and manipulated objects. By using five synchronized camera views, our reconstruction receives stronger geometric cues and produces more consistent novel-view renderings. We adopt a Gaussian Splatting-based approach for rendering realistic images while preserving 3D consistency. 
In contrast, diffusion-based novel-view generation methods may introduce visual artifacts, change object geometry, or generate images that appear synthetic in robot manipulation settings. 
Such methods may also require scene-specific fine-tuning, making them difficult to generalize across different tasks and environments. 
Our reconstruction-based pipeline avoids this issue by producing novel views directly from the observed multi-view scene. 

\subsection{VLA Policy Training}

After generating novel-view observations, we use the resulting dataset to train a Vision-Language-Action policy. 
The policy receives a visual observation and a natural language instruction as input and predicts the robot action required to complete the task. 
The key advantage of our new augmented dataset is that each generated observation preserves the original task state, allowing us to reuse the same language instruction and action label from the source demonstration.

Let \(\mathbb{D}\) denote the original demonstration dataset, and \(\mathbb{D}_\mathrm{nv}\) the dataset created through Instantsplat++ based novel-view rendering,
The final training dataset is defined as $\mathbb{D}_\mathrm{train} = \mathbb{D} \cup \mathbb{D}_\mathrm{nv}$. Each sample in \(\mathbb{D}_\mathrm{train}\) has the form \((o_t, l, a_t)\), where \(o_t\) may be an original observation, or a rendered novel view image.

We train the VLA policy using imitation learning. 
For continuous-action policies, the model is optimized to minimize the difference between the predicted action and the expert action: $\mathcal{L}_{\mathrm{act}}=\sum{(o_t,l,a_t)\in\mathbb{D}_\mathrm{train}}|\pi(o_t,l)-a_t|_2^2$. For action-token-based VLA models, the same training data can be used with a cross-entropy objective over discretized action tokens, $\mathcal{L}_{\mathrm{token}}=-\sum{(o_t,l,a_t)\in\mathbb{D}_\mathrm{train}}\log p{\pi}(a_t\mid o_t,l)$.

During training, each mini-batch contains a mixture of original and augmented observations. Since the same action may be paired with multiple visual variants of the same task state, the policy is encouraged to learn visual representations that are invariant to camera viewpoints. 
This reduces reliance on spurious cues such as fixed camera pose, tabletop color, or background texture, and encourages the model to focus on task-relevant information such as object identity, object location, gripper pose, and language-conditioned intent. Our augmentation framework is model-agnostic and can be applied to any VLA architecture. It does not require changing the policy architecture or action space; instead, it improves the visual diversity of the demonstration data used for training. At inference time, the policy receives only the current visual observation and language instruction, and no reconstruction module is required.

\subsection{Experimental setup}

We evaluate our approach on four manipulation tasks: \textit{Pick-and-Place Cube}, \textit{Sweep}, \textit{Sort}, and \textit{Stack}, chosen to span different levels of occlusion, object geometry, and manipulation precision. For each task, we collect 70 demonstrations using a synchronized five-camera setup mounted around an SO-101 manipulator. From these multi-view observations, we reconstruct a per-timestep 3D Gaussian representation using InstantSplat++.

We define camera azimuth with respect to the center of the manipulation workspace, with $0^\circ$ corresponding to the frontal viewing direction. The five physical cameras span approximately a $180^\circ$ arc around the workspace, and novel camera poses are restricted to $\theta \in \left[-{\pi}/{2}, {\pi}/{2}\right]$. We additionally vary the radial distance from the workspace center within $r \in [0.6, 1.2]\,\mathrm{m}$. Elevation is sampled from the table-plane level up to a top-down view, corresponding to $\phi \in \left[0, {\pi}/{2}\right]$. This sampling strategy avoids viewpoints behind the robot while introducing diversity in viewing angle, scale, and framing. We sample up to 15 novel third-person views per timestep. These views are then combined to construct up to 100 virtual episodes from each original demonstration, with camera combinations sampled independently across demonstrations.



We use SmolVLA as the policy architecture for all experiments, with three visual inputs: one wrist-mounted camera and two third-person cameras. For each virtual episode, we retain the original wrist-camera observations and action labels, while replacing the two third-person observations with synthesized views from our InfiNoVA pipeline. This results in 7,000 episodes for each of the tasks. The single-view baseline is trained using the original reference-camera observations, while the VISTA baseline is trained using data generated by VISTA. For a fair comparison, all policies use the same SmolVLA training configuration, are trained for the same number of epochs with a batch size of 64, and receive no method-specific hyperparameter tuning.

\section{results}

\subsection{Baselines}
For novel-view rendering, we compare InfiNoVA against methods spanning single-view generation, multi-view diffusion, explicit 3D generation, and Gaussian Splatting. \textbf{VISTA}~\cite{tian2025view} is the closest robotics-specific baseline, as it uses novel-view synthesis for viewpoint-robust policy augmentation; for a fair comparison, we fine-tune its NVS model on our environment. \textbf{GLD}~\cite{jang2026gld} provides a multi-view diffusion baseline, contrasting InfiNoVA's explicit Gaussian reconstruction with generative multi-view
synthesis. \textbf{Hunyuan3D}~\cite{hunyuan3d2025hunyuan3d} represents an explicit 3D-generation alternative, while \textbf{DepthSplat}~\cite{xu2025depthsplat} provides the closest reconstruction baseline by combining sparse multi-view inputs with feed-forward Gaussian Splatting.

For policy evaluation, we compare against a \textbf{Base Policy} (SmolVLA) trained on the original camera views and evaluated under unseen randomized viewpoints, and \textbf{VISTA-Aug}, trained with VISTA-generated views. We use VISTA as the policy-level synthesis baseline because it is the only competing method that remains sufficiently realistic and task-consistent for downstream augmentation; the others either exhibit substantial visual and geometric artifacts or incur impractical computational cost (Section~\ref{sec:rendering_time}).
\begin{figure*}[h]
\centering
\setlength{\tabcolsep}{2pt}
\renewcommand{\arraystretch}{1}
\resizebox{0.8\textwidth}{!}{%
\begin{tabular}{c c c c c}

 & \textbf{Pick-and-Place} & \textbf{Sweep} & \textbf{Stack} & \textbf{Sort} \\

\rowlabel{Hunyuan 3D}{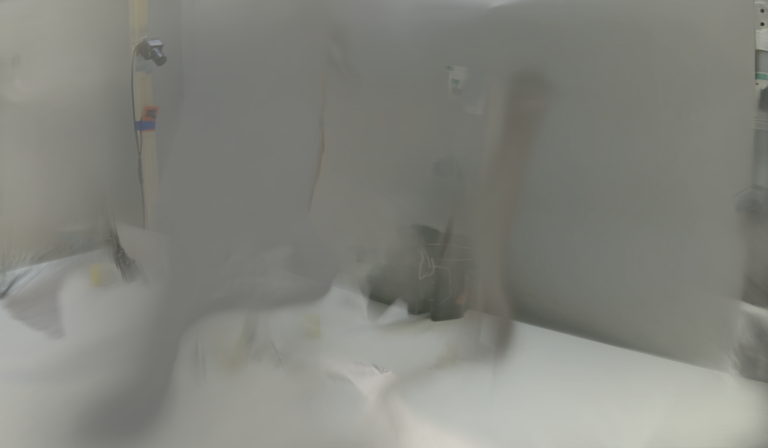} &
\includegraphics[width=0.2\linewidth]{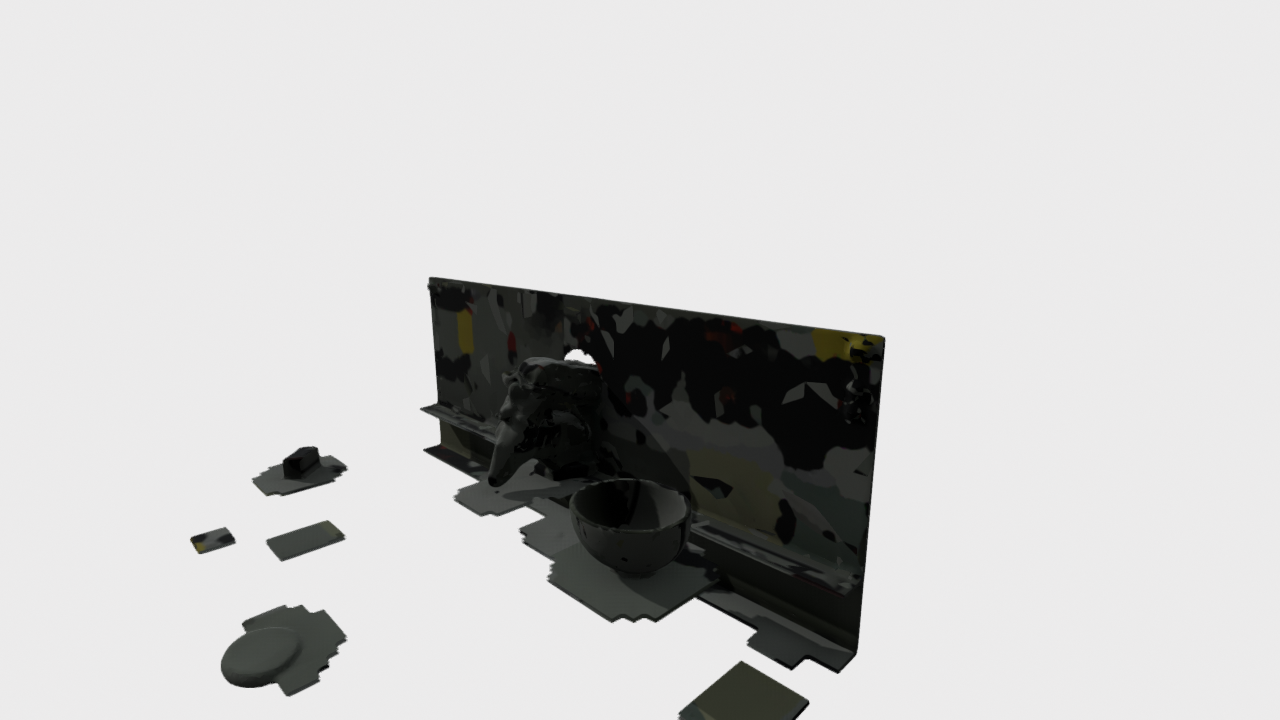} &
\includegraphics[width=0.2\linewidth]{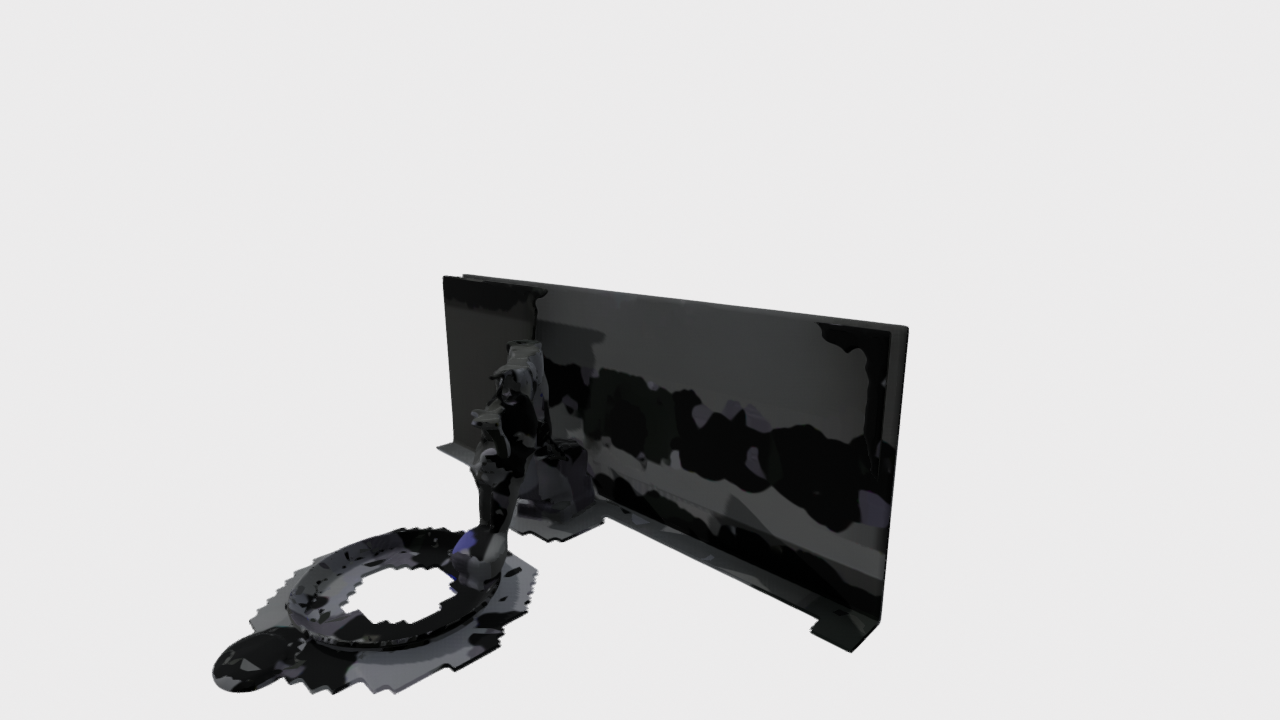}&
\includegraphics[width=0.2\linewidth]{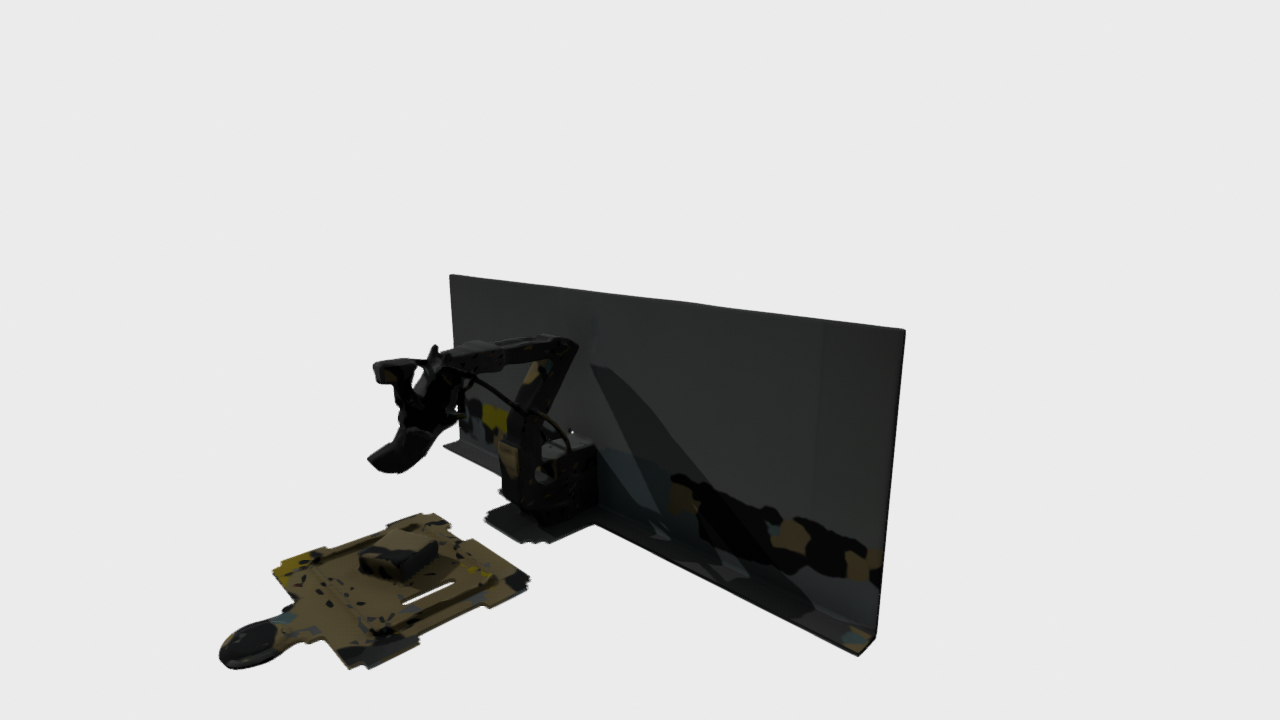} &
\includegraphics[width=0.2\linewidth]{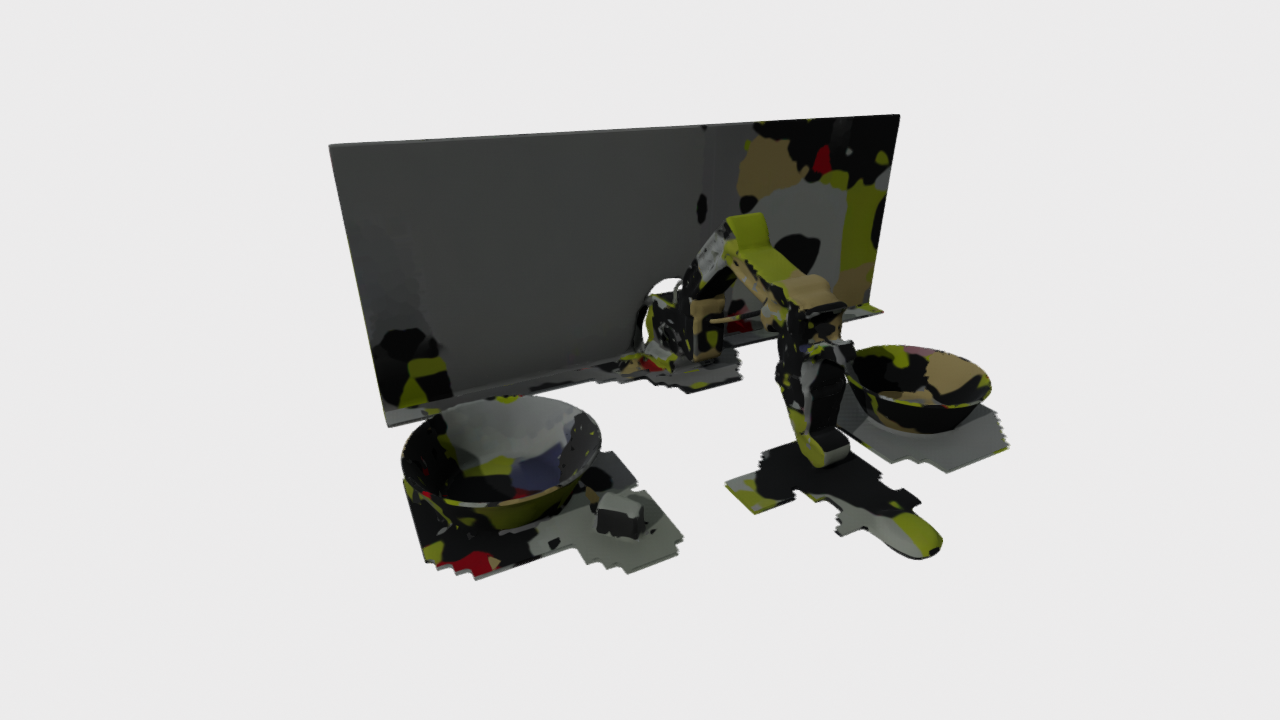} \\

\rowlabel{DepthSplat}{figures/depthsplat/pp_nv.png} &
\includegraphics[width=0.2\linewidth]{figures/depthsplat/pp_nv.png} &
\includegraphics[width=0.2\linewidth]{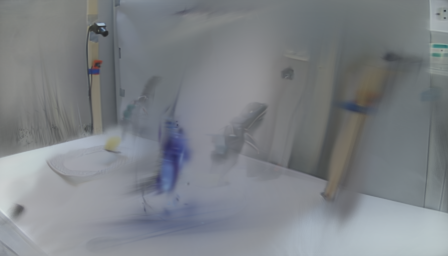} &
\includegraphics[width=0.2\linewidth]{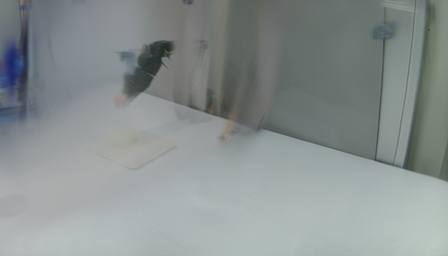} &
\includegraphics[width=0.2\linewidth]{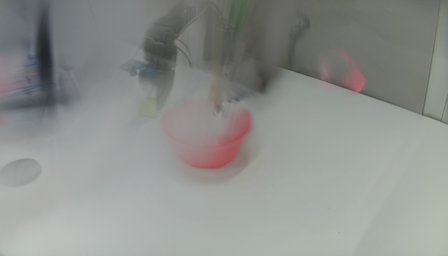} \\

\rowlabel{GLD}{figures/depthsplat/pp_nv.png} &
\includegraphics[width=0.2\linewidth]{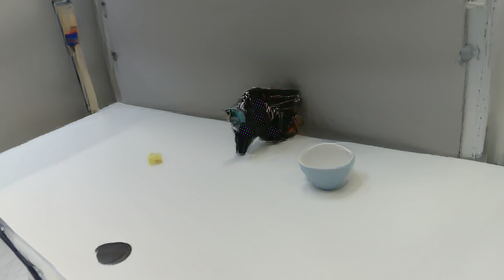} &
\includegraphics[width=0.2\linewidth]{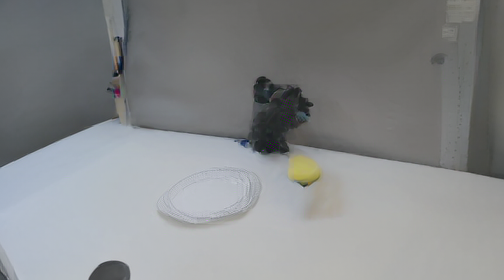} &
\includegraphics[width=0.2\linewidth]{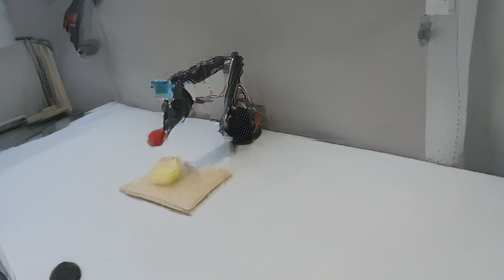} &
\includegraphics[width=0.2\linewidth]{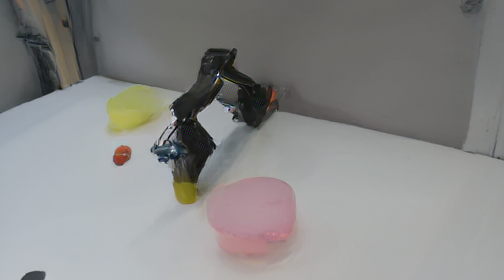} \\

\rowlabel{VISTA}{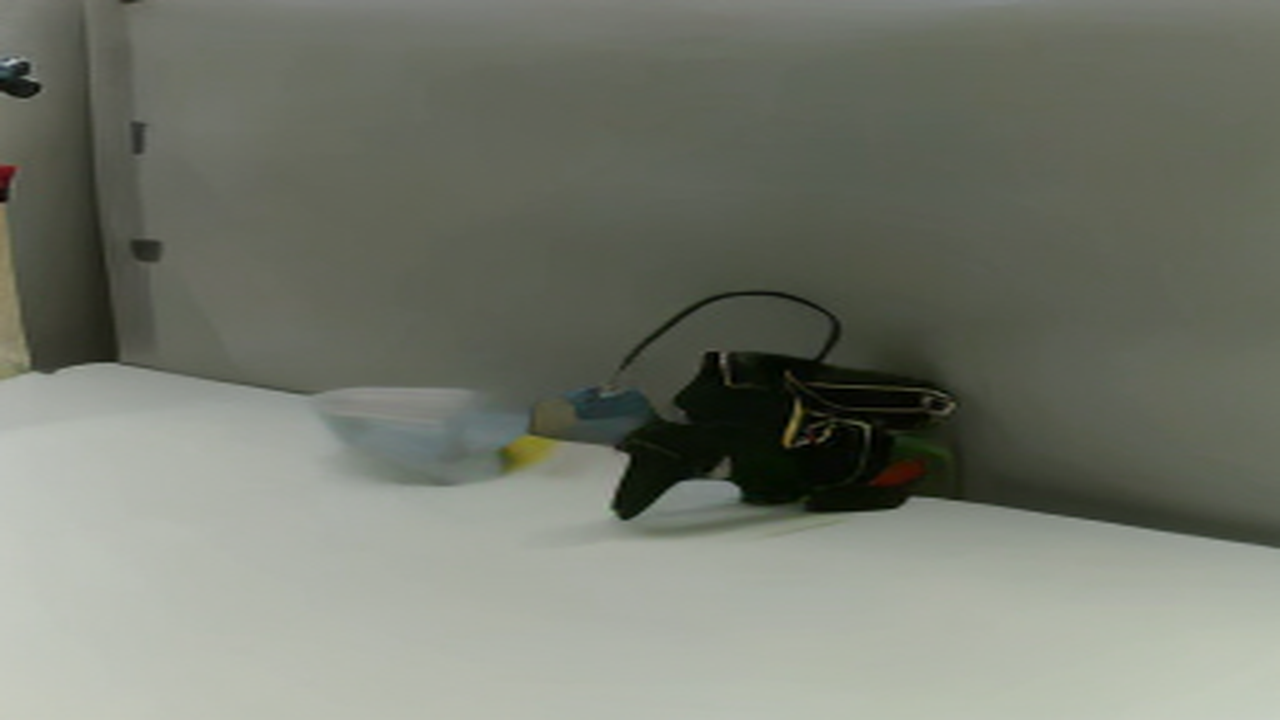} &
\errorellipse{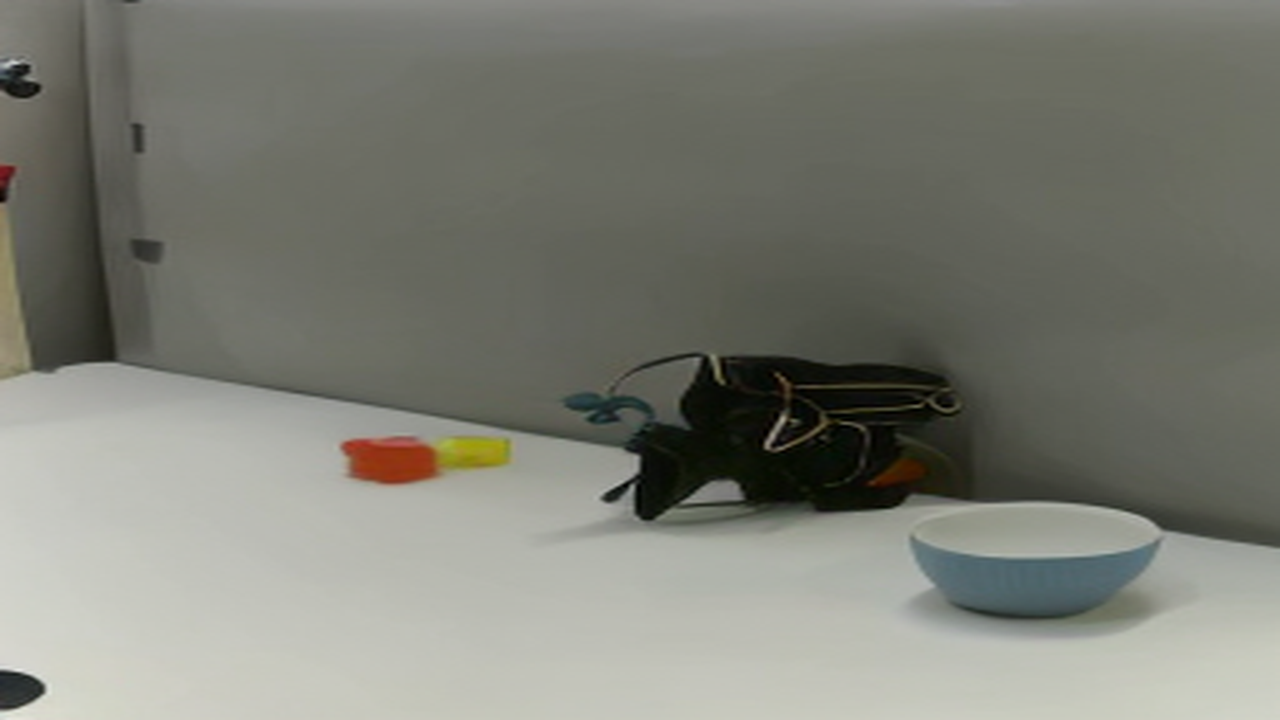}{0.325}{0.375}{0.10}{0.10}{0} &
\errorellipse{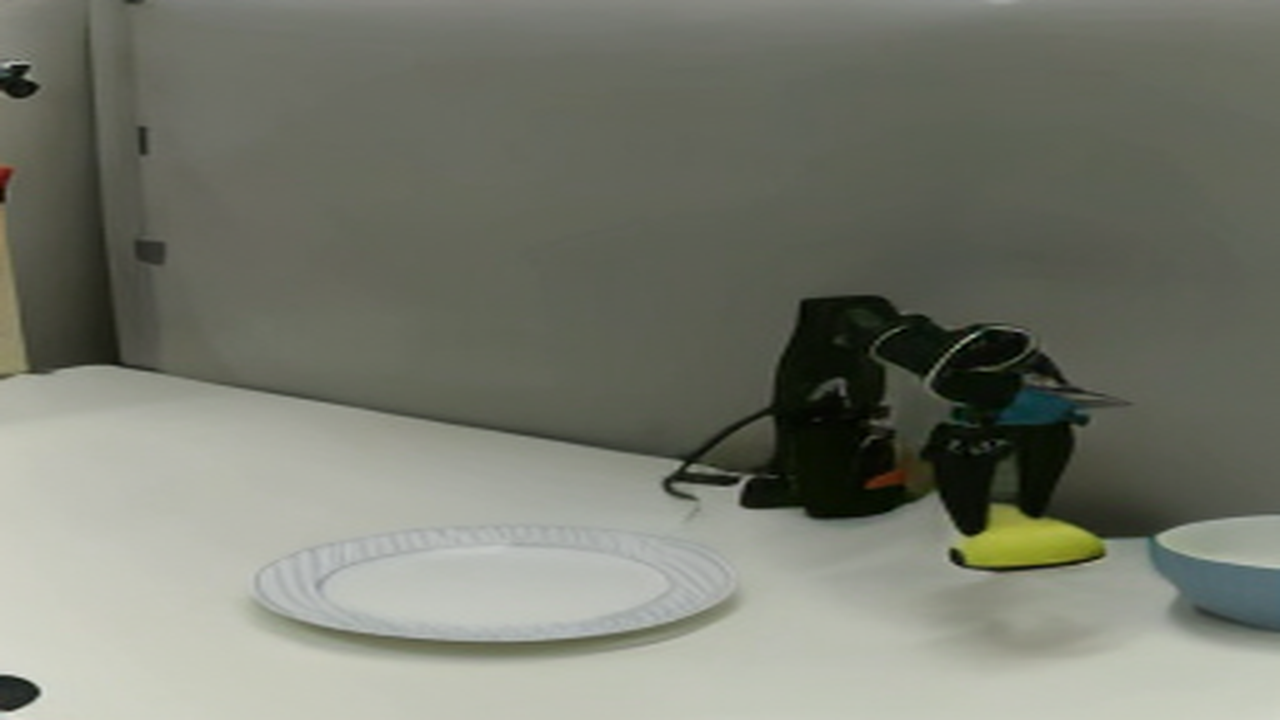}{0.945}{0.225}{0.08}{0.08}{0} &
\errorellipse{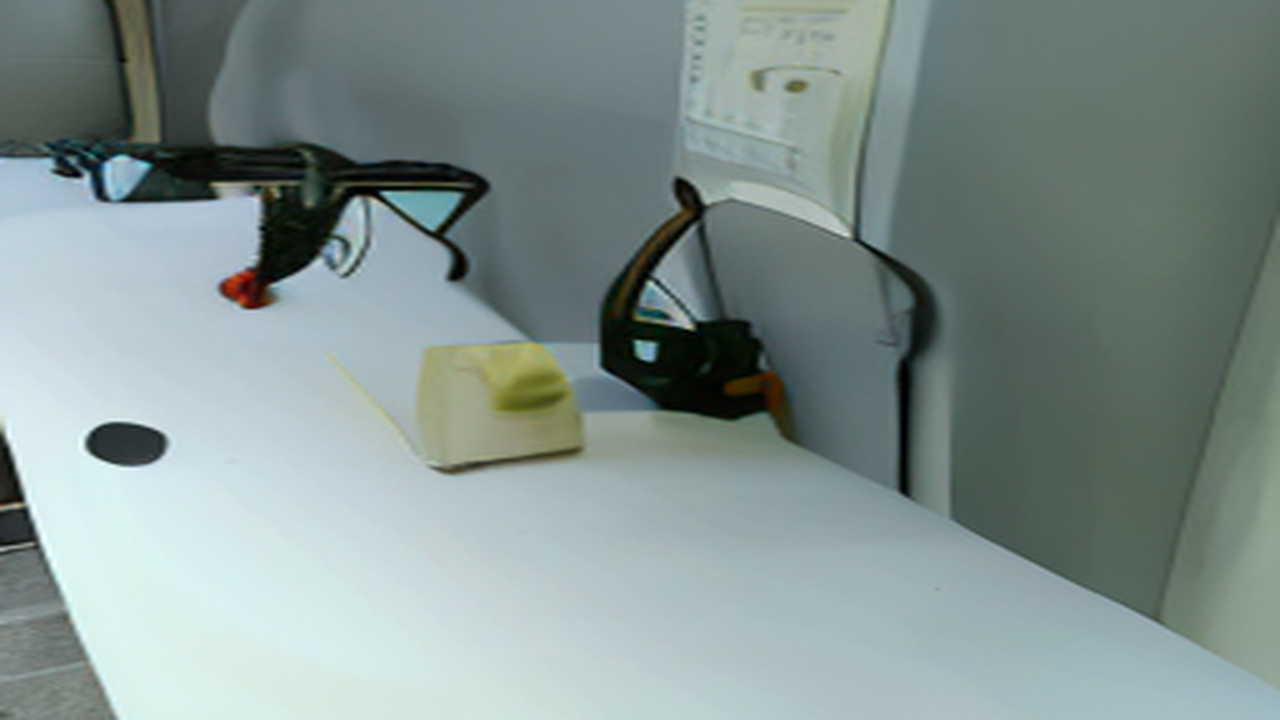}{0.45}{0.75}{0.125}{0.10}{0} &
\errorellipse{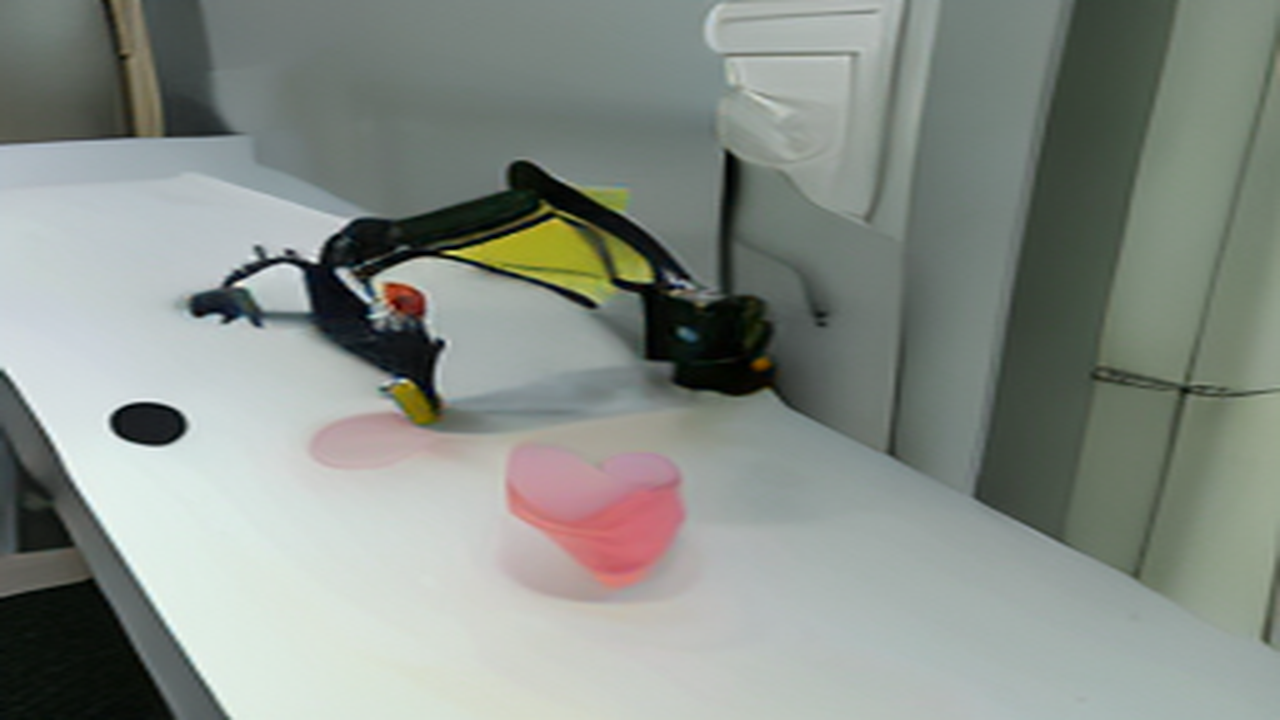}{0.35}{0.70}{0.15}{0.10}{25} \\

\rowlabel{Ours}{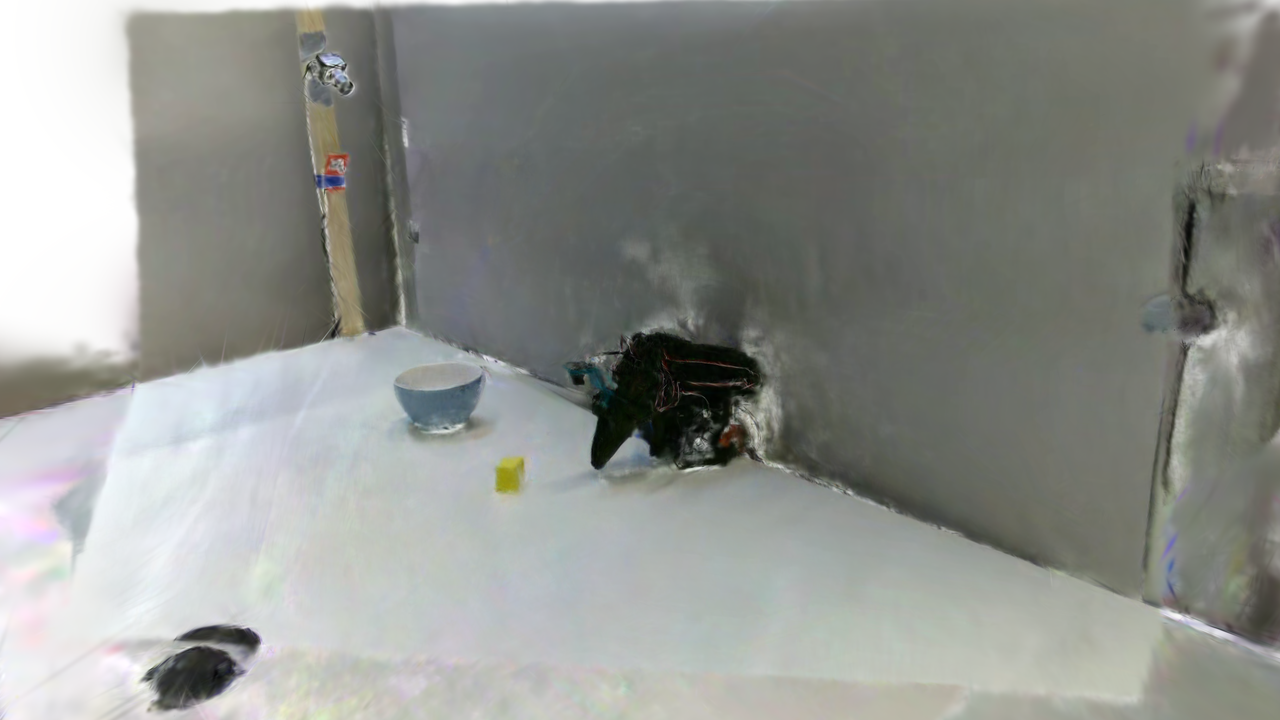} &
\includegraphics[width=0.2\linewidth]{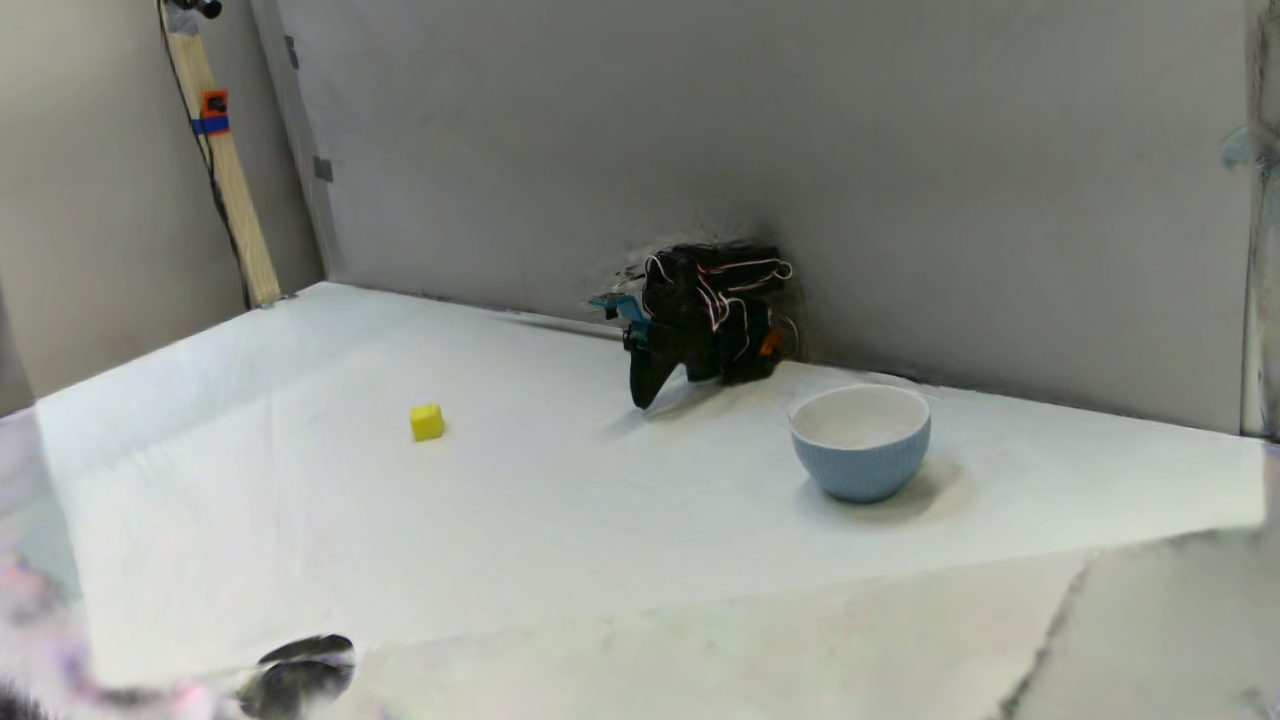} &
\includegraphics[width=0.2\linewidth]{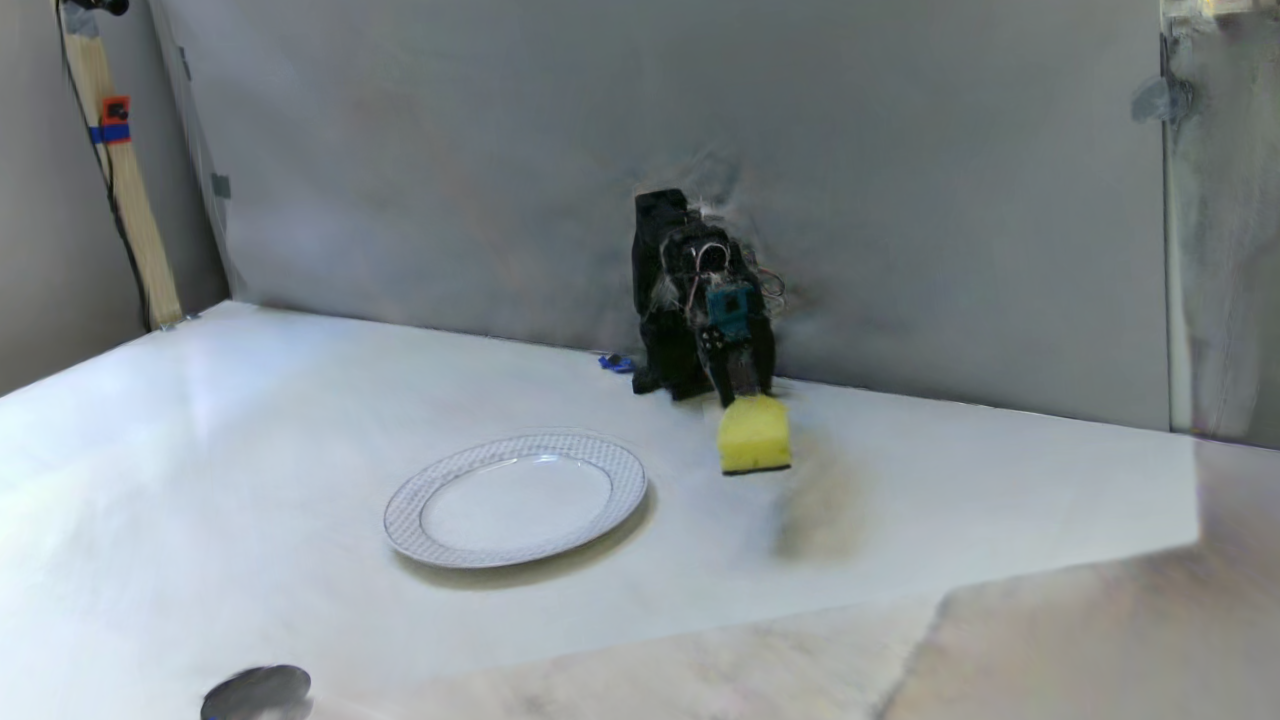} &
\includegraphics[width=0.2\linewidth]{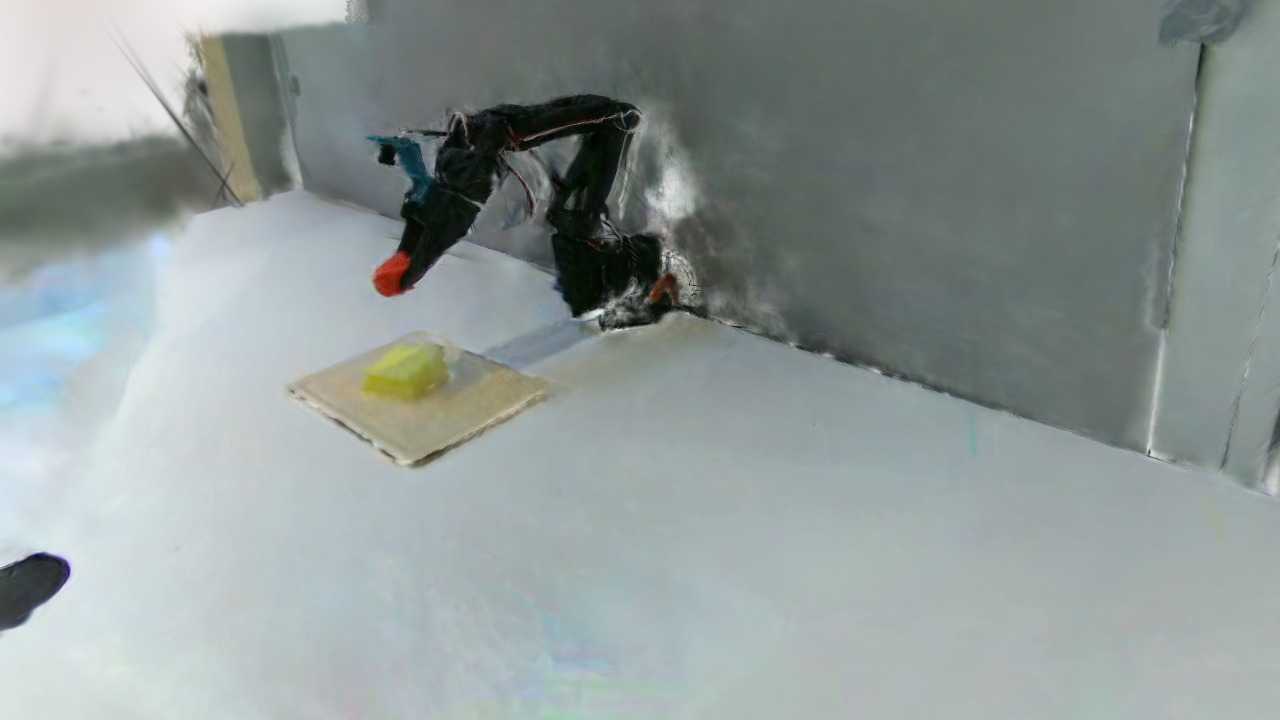} &
\includegraphics[width=0.2\linewidth]{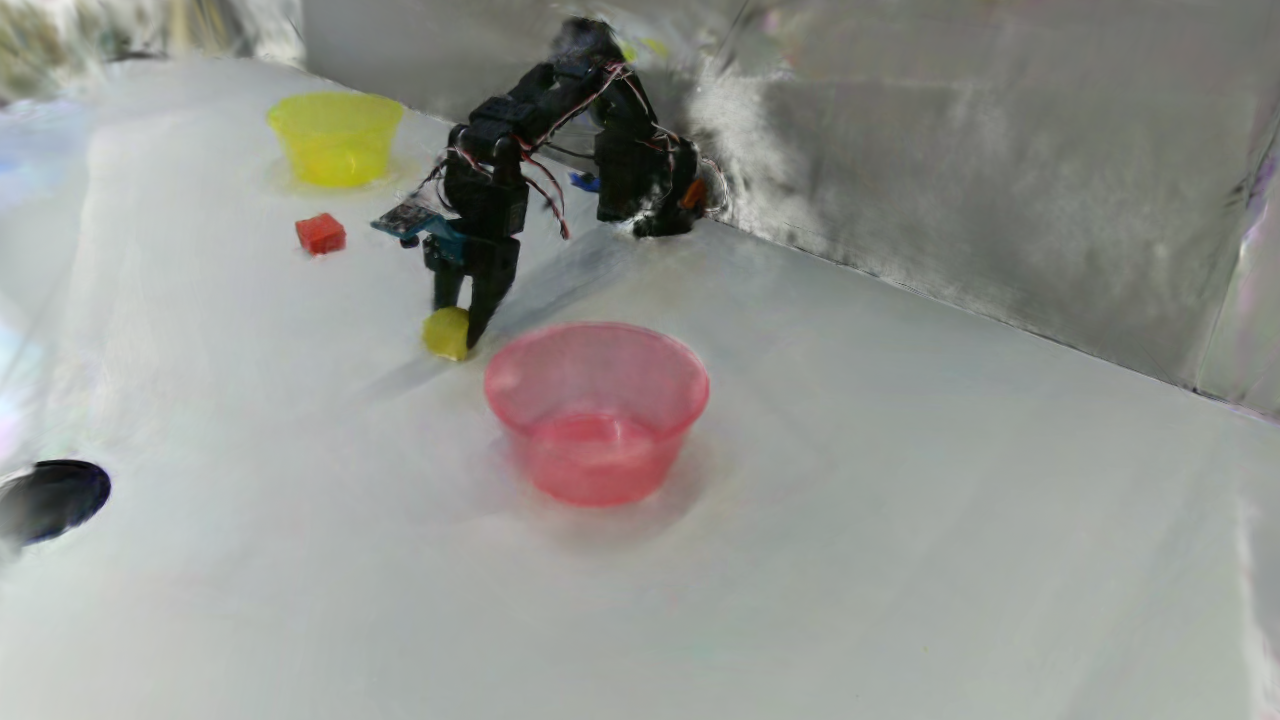} \\

\rowlabel{GT}{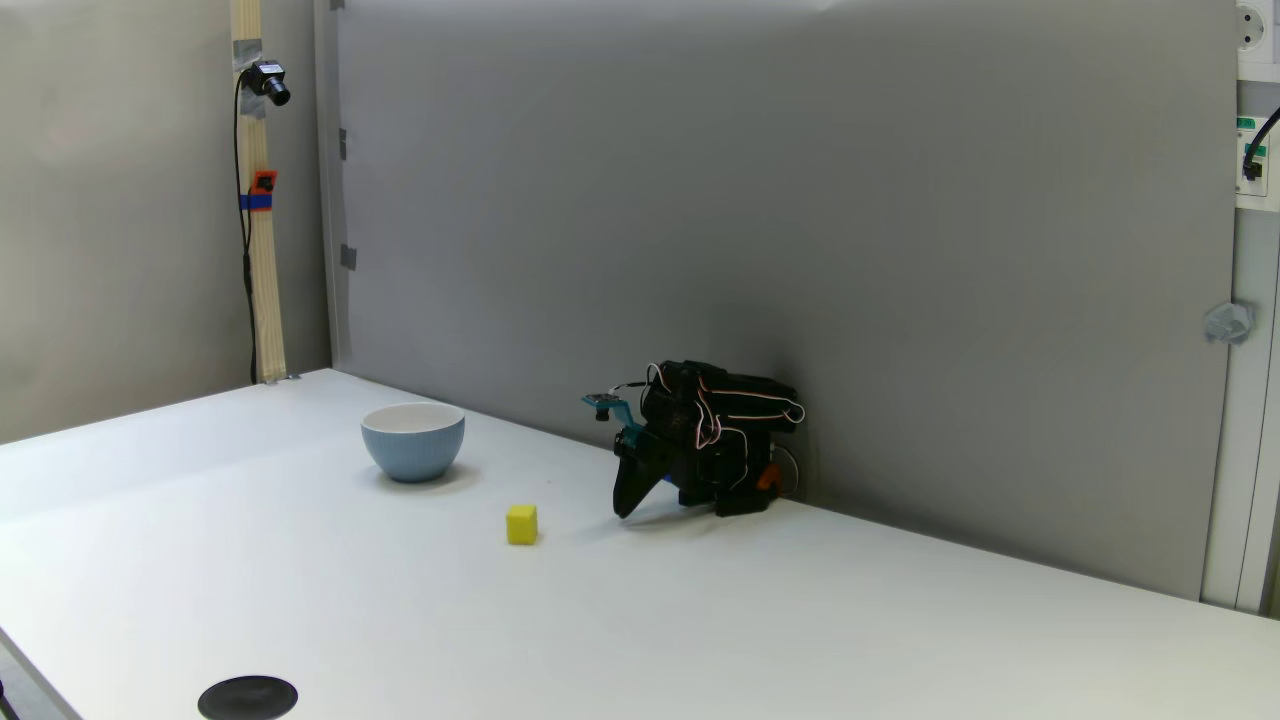} &
\includegraphics[width=0.2\linewidth]{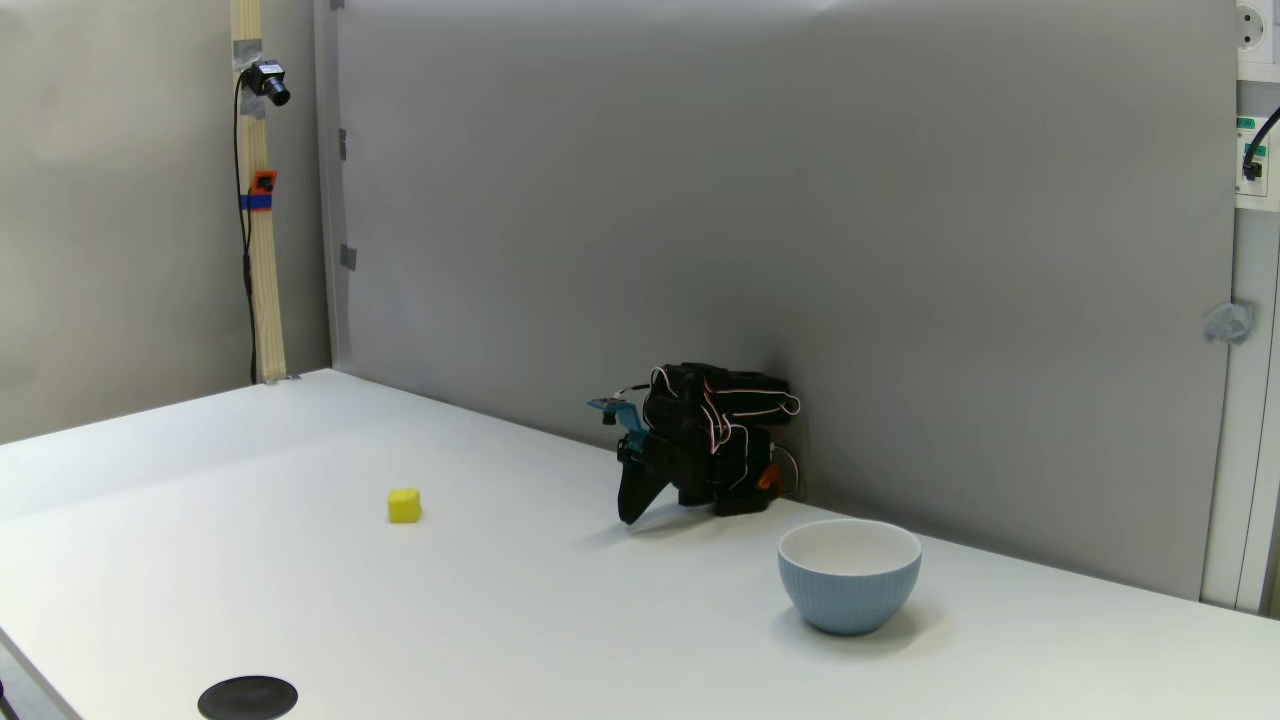} &
\includegraphics[width=0.2\linewidth]{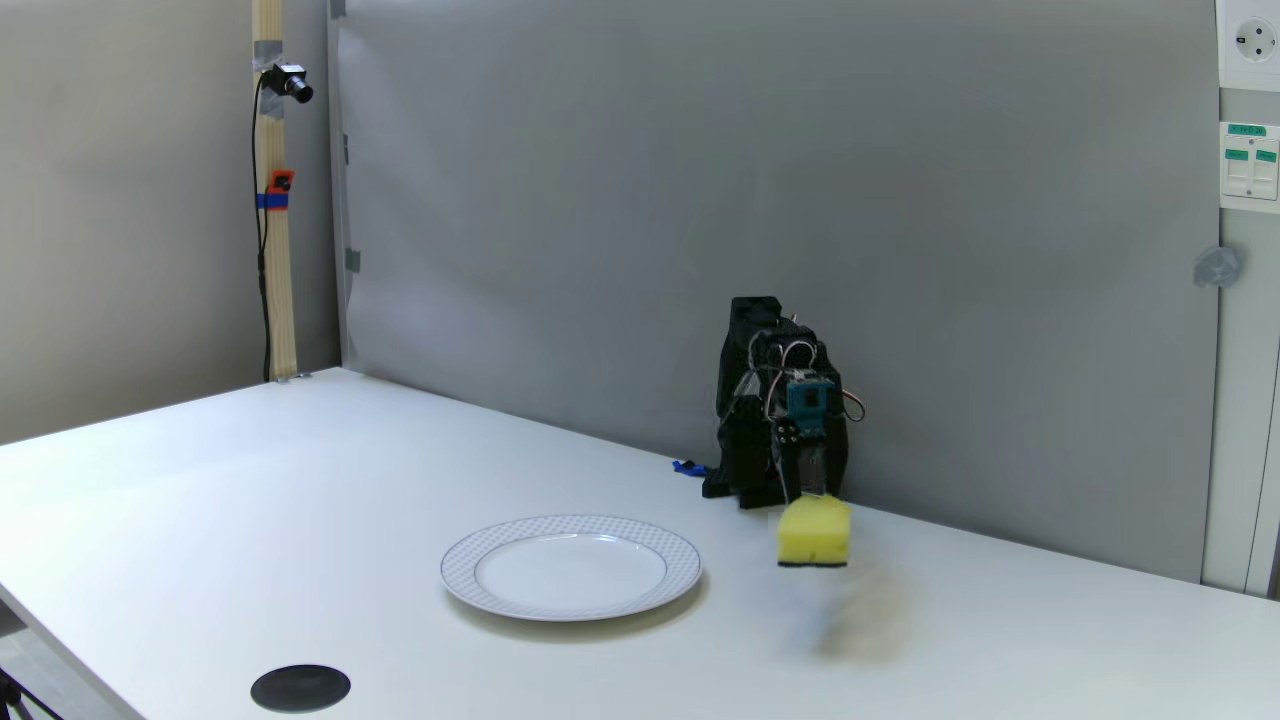} &
\includegraphics[width=0.2\linewidth]{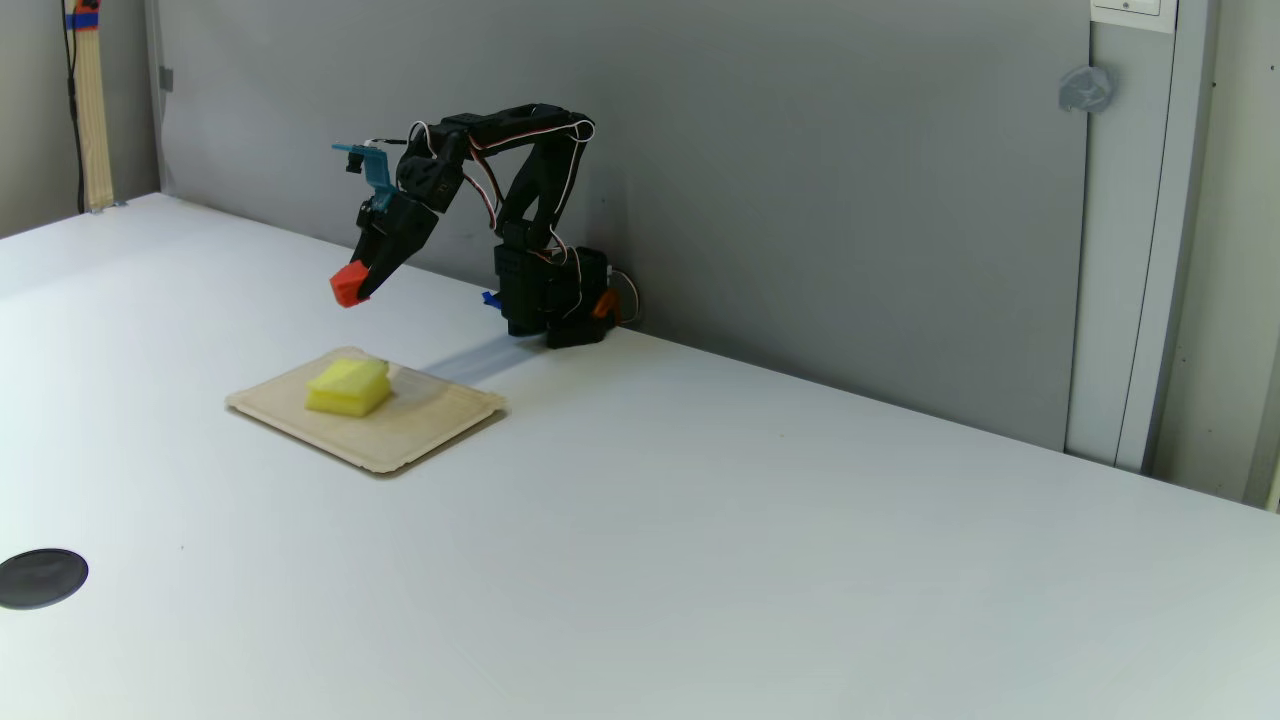} &
\includegraphics[width=0.2\linewidth]{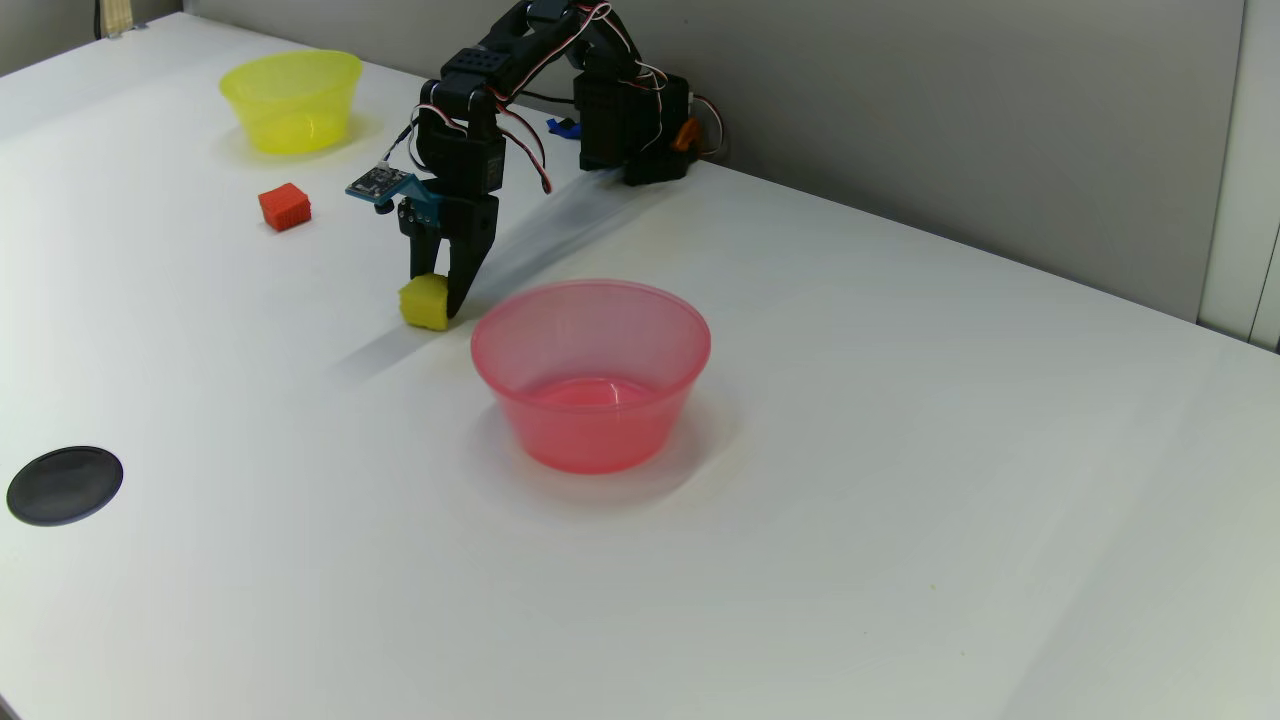} \\

\end{tabular}%
}

\caption{Qualitative comparison of novel-view synthesis across the four manipulation tasks. Rows correspond to competing synthesis methods. Red ellipses highlight representative task-critical artifacts in VISTA.}
\label{fig:NVS}
\vspace{-0.5cm}
\end{figure*}

\subsection{Quality of Novel-view Renderings}
\label{sec:rendering_quality}
We first evaluate whether multi-view 3D reconstruction provides higher-quality and more reliable novel-view augmentation than existing synthesis approaches (Fig.~\ref{fig:NVS}). We compare all methods using the same source observations and target viewpoints, randomly sampling 100 observations from each of the four tasks, for a total of 400 evaluation samples. Each generated view is compared against the corresponding real camera observation using PSNR, SSIM~\cite{wang2004image}, and LPIPS~\cite{zhang2018unreasonable}. As reported in Table~\ref{tab:nvs_quality}, InfiNoVA achieves the best performance across all three metrics, with a PSNR of 16.65, SSIM of 0.841, and LPIPS of 0.388. Compared with the three strongest baselines, InfiNoVA improves PSNR by 4.3--25.5\% and SSIM by 2.6--11.7\%, while reducing LPIPS by 15.5--25.0\%. These results demonstrate consistently higher pixel-level, structural, and perceptual fidelity across the evaluated novel views.

\vspace{-0.15cm}
\begin{table}[h]
\centering
\caption{Quantitative comparison of novel-view rendering quality. Higher PSNR and SSIM indicate better reconstruction fidelity, while lower LPIPS indicates better perceptual similarity.}
\label{tab:nvs_quality}
\footnotesize
\begin{tabular*}{\columnwidth}{@{\extracolsep{\fill}}lccc@{}}
\toprule
Method & PSNR $\uparrow$ & SSIM $\uparrow$ & LPIPS $\downarrow$ \\
\midrule
Hunyuan 3D~\cite{hunyuan3d2025hunyuan3d} & 7.48 & 0.693 & 0.596 \\
DepthSplat~\cite{xu2025depthsplat} & 13.27 & 0.809 & 0.517 \\
GLD~\cite{jang2026gld} & 15.97 & 0.753 & 0.485 \\
VISTA~\cite{tian2025view} & 15.13 & 0.820 & 0.459 \\
\textbf{InfiNoVA (Ours)} & \textbf{16.65} & \textbf{0.841} & \textbf{0.388} \\
\bottomrule
\end{tabular*}
\end{table}
\vspace{0.1cm}

The qualitative results in Fig.~\ref{fig:NVS} further expose the main failure modes of the competing methods. Hunyuan3D often reconstructs incomplete workspaces with fragmented objects, while DepthSplat preserves the coarse layout but produces blurred robot and object geometry. GLD gives the strongest visual quality among the baselines, but incurs substantially higher computational cost (Table~\ref{tab:nvs_time}).
VISTA generates realistic-looking images but introduces task-critical hallucinations, highlighted by the red ellipses in Fig.~\ref{fig:NVS}. In Pick-and-Place, it incorrectly introduces two sponges that belong to the Stack task, while in Sweep it hallucinates the bowl observed in Pick-and-Place. In Stack, part of the robot arm is missing, and in Sort the arm geometry is visibly distorted. 

Because VISTA is fine-tuned on the robot-task data but is not explicitly constrained by the task identity or complete 3D state of the current episode, ambiguous or disoccluded regions can be filled using dataset-level visual priors, leading to cross-task content leakage. This is particularly problematic for policy augmentation because the hallucinated frame retains the original action label. In contrast, InfiNoVA renders from an explicit 3D Gaussian Splat reconstructed from synchronized multi-view observations of the current scene, so object configurations andndisoccluded regions are grounded in observed geometry, preserving the robot, manipulated objects, and scene structure more reliably and without borrowing task-specific content from other demonstrations.

\subsection{Rendering Time}
\label{sec:rendering_time}
Table~\ref{tab:nvs_time} compares the computational cost of novel-view synthesis, including reconstruction and rendering, for the strongest methods from Section~\ref{sec:rendering_quality}. Although GLD produces competitive renderings, its computational cost is prohibitively high, making trajectory-scale generation impractical for temporal and downstream policy evaluation; we therefore exclude it
from those experiments. Averaged across tasks, InfiNoVA is approximately $3.6\times$ faster than VISTA and $82.8\times$ faster than GLD, while also providing higher rendering fidelity and temporal consistency (Section~\ref{sec:rendering_quality}). Overall, InfiNoVA achieves the strongest quality--efficiency trade-off among the evaluated methods.
\vspace{-0.1cm}

\begin{table}[h]
\centering
\caption{Novel-view synthesis time comparison (in seconds).}
\label{tab:nvs_time}
\footnotesize
\setlength{\tabcolsep}{2pt}
\begin{tabular*}{\columnwidth}
{@{\extracolsep{\fill}}lccc@{}}
\toprule
Task & GLD & VISTA & InfiNoVA \\
\midrule
Pick-and-Place  & 25615 & 1115 & 318 \\
Sweep           & 35119 & 1529 & 437 \\
Sort            & 39026 & 1693 & 434 \\
Stack           & 45056 & 1961 & 560 \\
\midrule
Average/episode & 36204 & 1575 & 437 \\
\bottomrule
\end{tabular*}
\vspace{-0.2cm}
\end{table}

\subsection{Temporal and Geometric Consistency}
We next evaluate whether the generated trajectories preserve the temporal evolution and task state of the original demonstrations. Beyond frame-wise rendering quality, a valid augmentation should reproduce the motion observed in the real trajectory without introducing artificial changes across consecutive frames. To quantify this, we randomly sample 10 trajectories from each of the four tasks, and evaluate them using two complementary feature-space temporal metrics. 1)~\textit{FeaCD}~\cite{gong2025temcoco} evaluates ResNet-18~\cite{he2016deep} features from consecutive frames and their inter-frame feature differences are treated as temporal change vectors, and cosine distance measures whether the generated and ground-truth trajectories evolve in the same direction in feature space. 2) We additionally introduce \textit{ReStraV-$\Delta\theta$}, a reference-based adaptation of ReStraV~\cite{interno2026ai}. ReStraV represents video frames using DINOv2~\cite{oquab2024dinov2learningrobustvisual} features and characterizes temporal trajectories through their local curvature. Given frame representations $z_t$, we first compute consecutive displacements $\Delta z_t = z_{t+1}-z_t$ and the local trajectory angle
\begin{equation}
\theta_t =
\arccos
\left(
\frac{\Delta z_t^\top \Delta z_{t+1}}
{\|\Delta z_t\|_2\|\Delta z_{t+1}\|_2}
\right).
\end{equation}
Rather than using curvature to distinguish real and generated videos as in the
original formulation, we compute these angles independently for the synthesized
and ground-truth trajectories and report their mean absolute angular deviation,
\begin{equation}
ReStraV-\Delta\theta =
\frac{1}{T-2}
\sum_{t=1}^{T-2}
\left|
\hat{\theta}_t-\theta_t^{\mathrm{GT}}
\right|.
\end{equation}
This measures whether the synthesized trajectory follows the same local geometry through DINOv2 feature space as the real trajectory. As reported in
Table~\ref{tab:temporal_consistency}, InfiNoVA achieves the best performance on both metrics, with a FeaCD of $0.915$ and a ReStraV-$\Delta\theta$ of
$3.49^\circ$. Compared with the next-best baseline, DepthSplat, this corresponds to reductions of $5.6\%$ in FeaCD and $32.8\%$ in angular deviation. Relative to VISTA, the reductions are $7.0\%$ and $51.3\%$. These gains show that InfiNoVA better preserves both the frame-to-frame direction of semantic change and the higher-order temporal geometry of the real trajectory, yielding a more coherent progression of the underlying manipulation state.
\begin{table}[ht]
\centering
\caption{Temporal consistency averaged across the four manipulation tasks.
Lower values indicate better agreement with the ground-truth trajectories.}
\label{tab:temporal_consistency}
\begin{tabular*}{\columnwidth}{@{\extracolsep{\fill}}lcc@{}}
\toprule
Method & FeaCD $\downarrow$ & ReStraV-$\Delta\theta$ ($^\circ$) $\downarrow$ \\
\midrule
Hunyuan 3D~\cite{hunyuan3d2025hunyuan3d}
    & 0.9780 & 6.09 \\
DepthSplat~\cite{xu2025depthsplat}
    & 0.969 & 5.19 \\
VISTA~\cite{tian2025view}
    & 0.984 & 7.16 \\
\textbf{InfiNoVA (Ours)}
    & \textbf{0.915} & \textbf{3.49} \\
\bottomrule
\end{tabular*}
\end{table}
    
\begin{figure}[h]
    \centering
    \setlength{\tabcolsep}{2pt}
    \begin{tabular}{c c c}
        & \textbf{Ours} & \textbf{VISTA} \\[3pt]
        \raisebox{1.6\height}{\rotatebox{90}{\textbf{$t_{90}$}}} &
        \includegraphics[width=0.42\columnwidth]{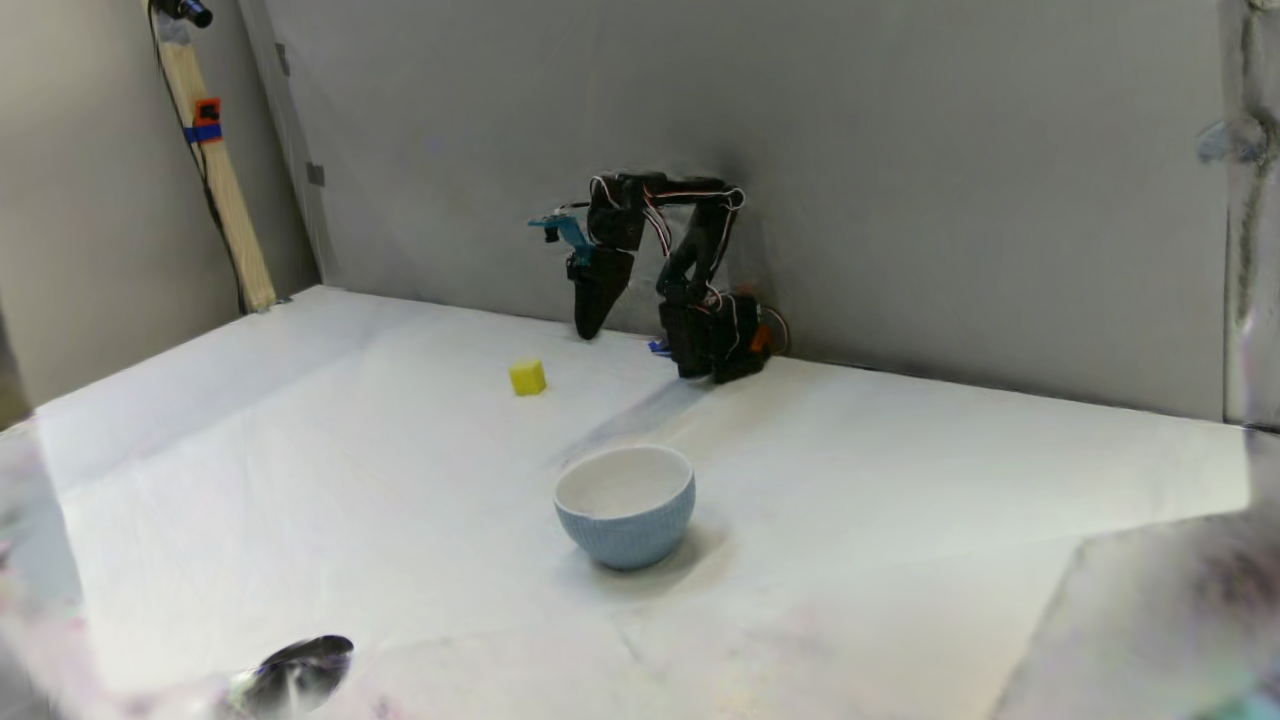} &
        \errorellipsevar{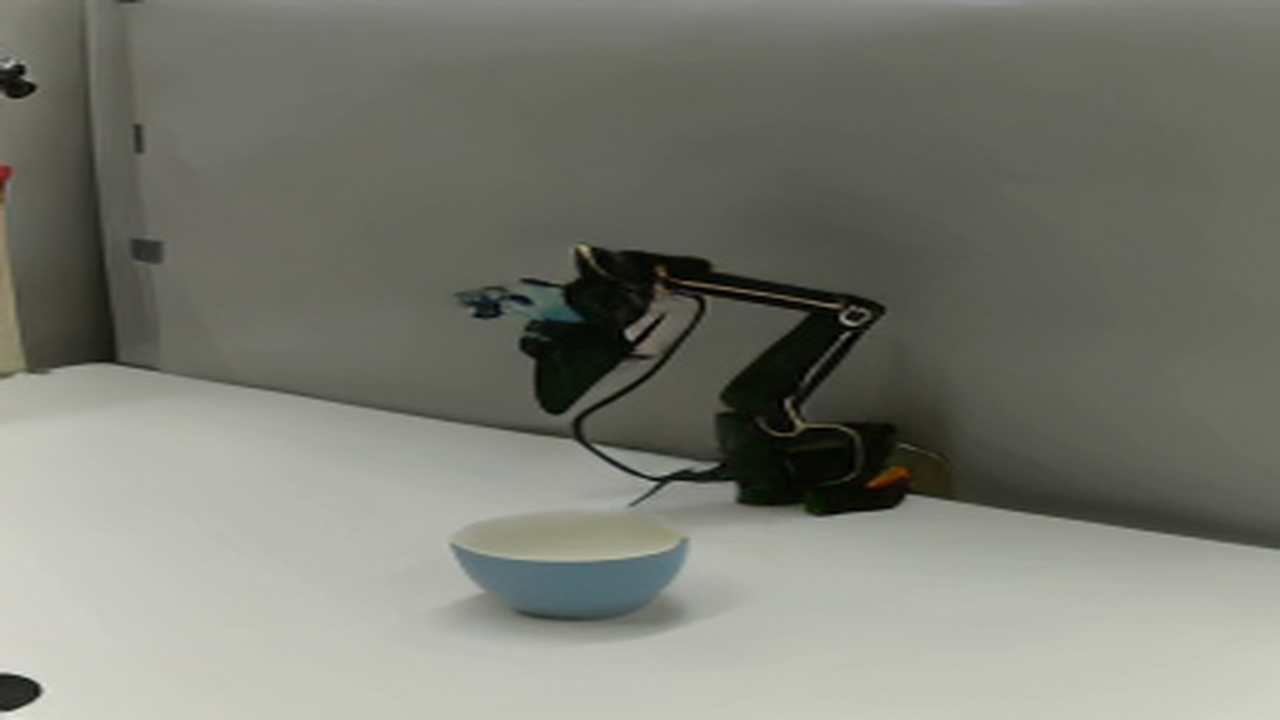}{0.350}{0.350}{0.05}{0.05}{0}{0.42}
        \\[4pt]

        \raisebox{1.4\height}{\rotatebox{90}{\textbf{$t_{494}$}}} &
        \includegraphics[width=0.42\columnwidth]{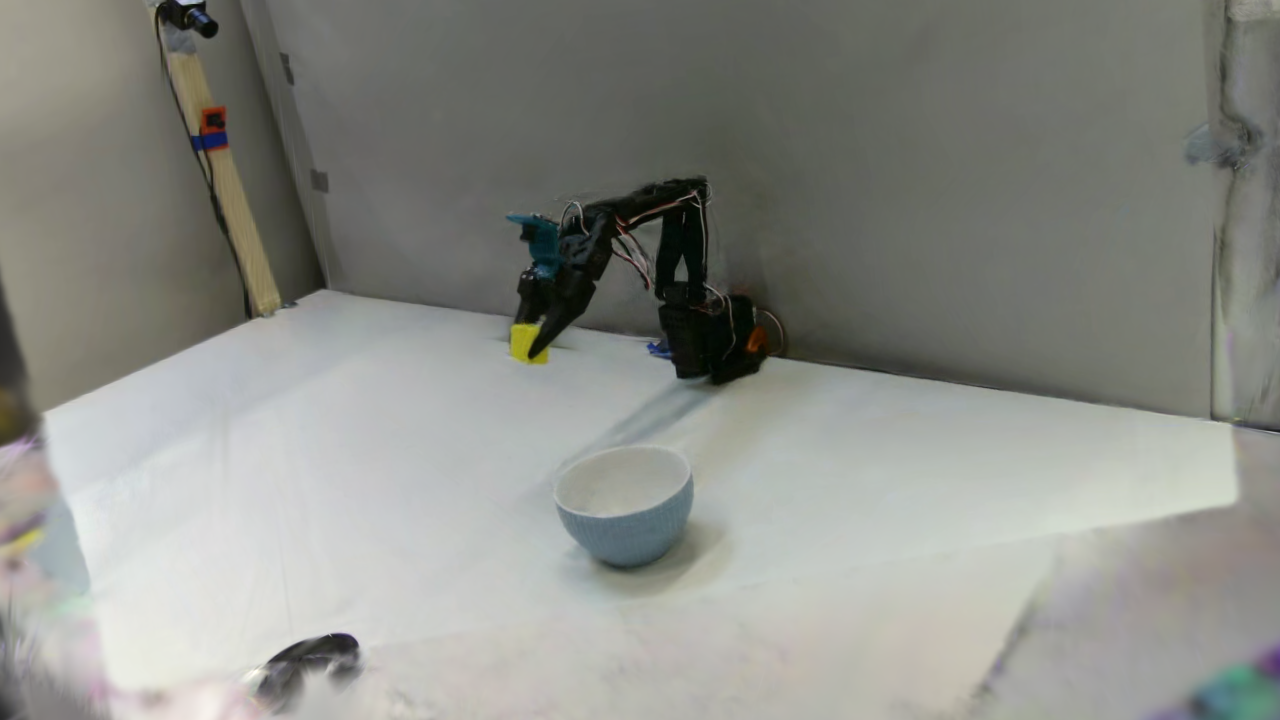} &
        \errorellipsevar{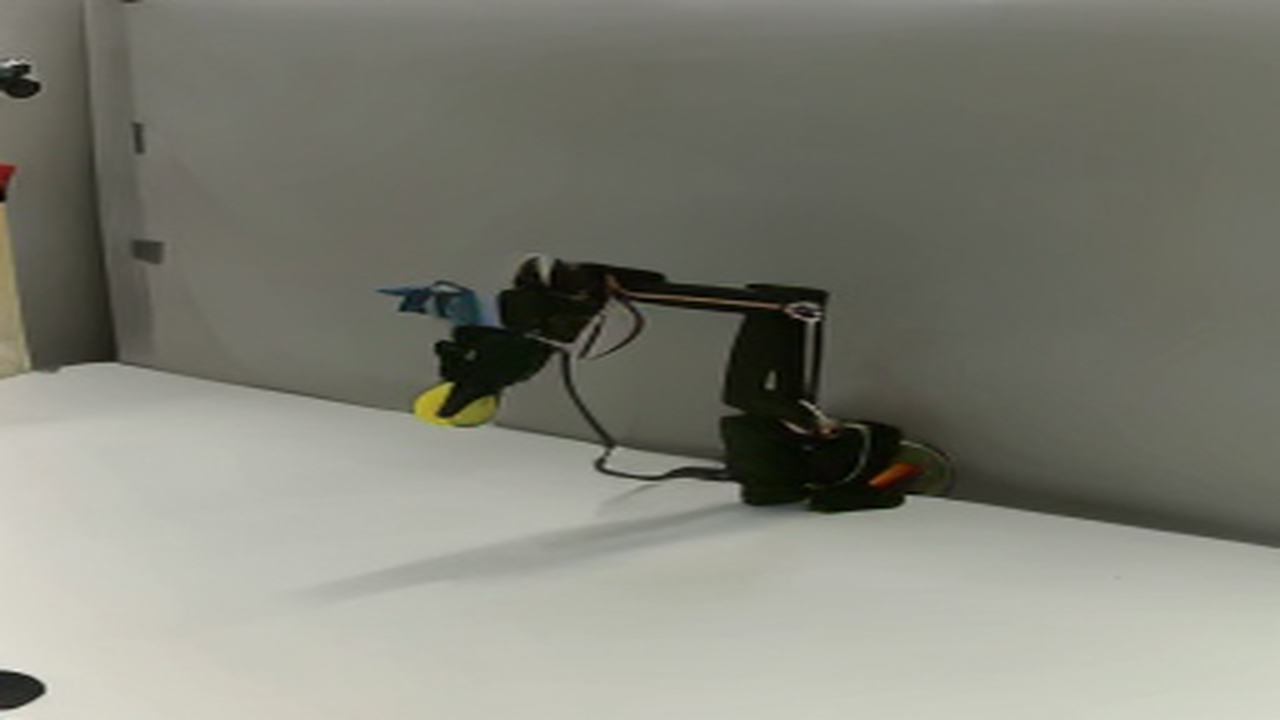}{0.6}{0.15}{0.075}{0.075}{0}{0.42}
        \\[4pt]

        \raisebox{1.4\height}{\rotatebox{90}{\textbf{$t_{512}$}}} &
        \includegraphics[width=0.42\columnwidth]{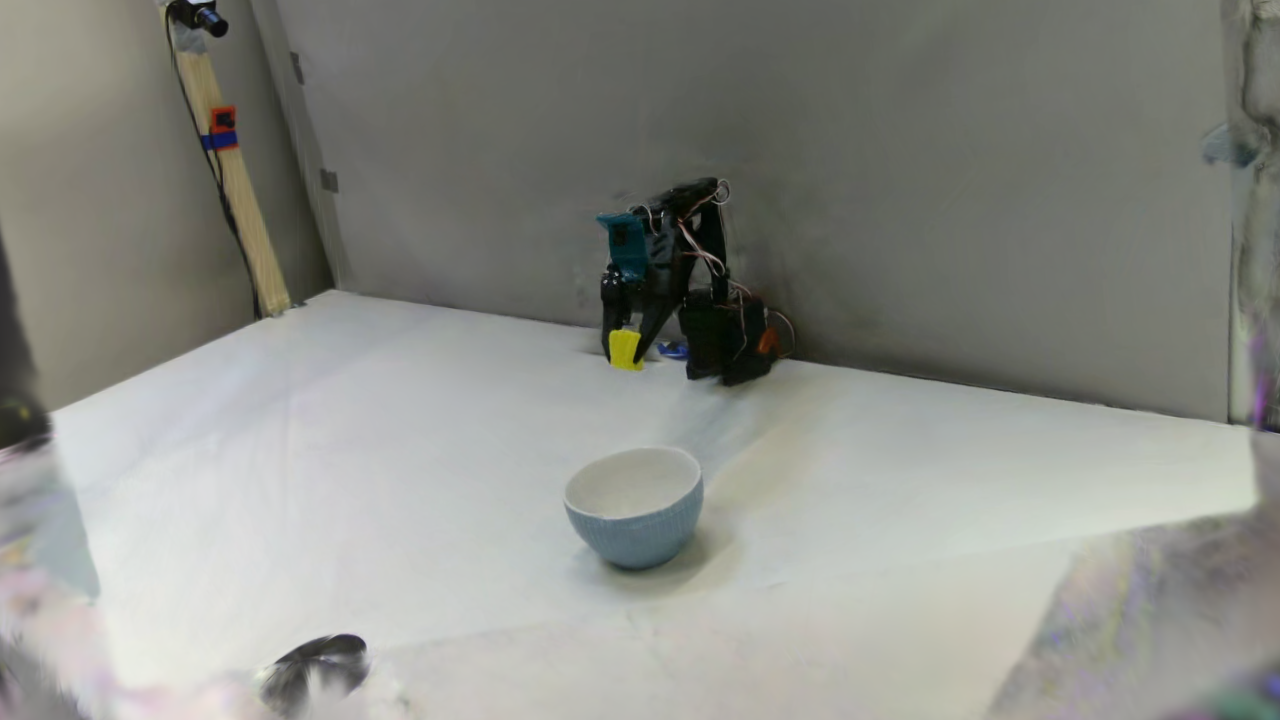} &
        \errorellipsevar{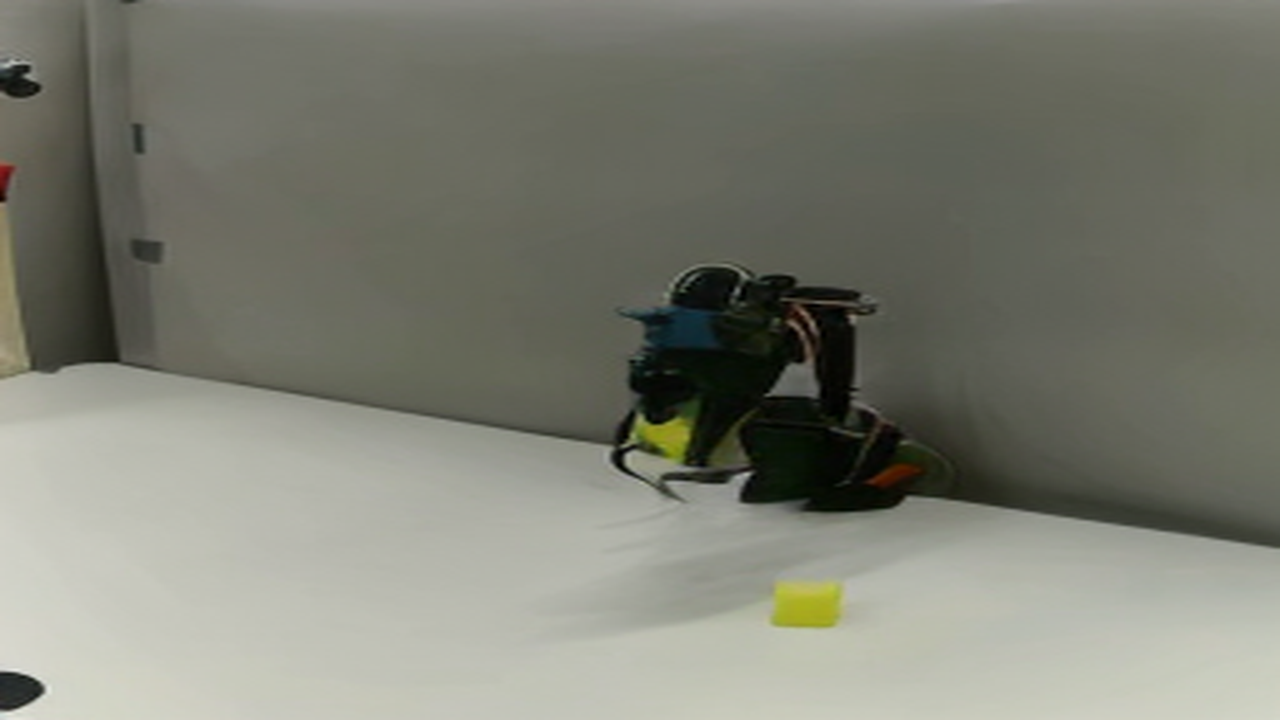}{0.625}{0.175}{0.065}{0.065}{0}{0.42}
    \end{tabular}
    \caption{Temporal consistency across three timesteps ($t_{90}$, $t_{494}$, and $t_{512}$) sampled from the same Pick-and-Place episode. Red ellipses highlight VISTA's temporal inconsistencies.}
    \label{fig:temporal_consistency}
    \vspace{-0.5cm}
\end{figure}

    The qualitative results in Fig.~\ref{fig:temporal_consistency} further expose these temporal failures. The three observations are sampled from the same Pick-and-Place episode at timestep $t_{90}$, $t_{494}$, and $t_{512}$. At $t_{90}$, VISTA completely removes the yellow cube that is present in the corresponding scene. At $t_{494}$, the bowl disappears despite remaining part of the same episode. More severely, at $t_{512}$ a yellow cube appears at the location of the missing bowl while the robot is simultaneously holding another cube, resulting in two instances of an object that should occur only once. Such inconsistencies are particularly harmful for policy augmentation because each synthesized observation retains the action label of the corresponding real state. Removing, duplicating, or replacing task-relevant objects therefore pairs the expert action with an observation describing a different physical state, introducing contradictory state--action supervision. InfiNoVA instead preserves these scene elements across the sampled timesteps, remaining consistent with the progression of the real episode.

\subsection{Rendering Precision}
We next evaluate whether novel-view synthesis produces repeatable outputs under identical input conditions for three best methods based on their rendering quality. Repeatability is relevant for policy training because each synthesized observation retains the original action label, while uncontrolled changes in robot or object state can weaken state--action correspondence. We sample $N=5$ observations and generate each target view $K=5$ times, keeping the source images and target camera pose fixed while varying the random seed. For observation $n$, let $\hat{I}_n^{(k)}=G(x_n,c_n;\xi_k)$ denote the output generated from source observations $x_n$, target camera pose $c_n$, and random
seed $\xi_k$. Following the LPIPS-based diversity measure of~\cite{ojha2021few}, we quantify rendering precision using the average pairwise perceptual variation,
\begin{equation}
\label{eq:rendering_precision}
V_{\mathrm{LPIPS}} =
\frac{2}{NK(K-1)}
\sum_{n=1}^{N}\sum_{i=1}^{K-1}\sum_{j=i+1}^{K}
\operatorname{LPIPS}(\hat{I}_n^{(i)},\hat{I}_n^{(j)}),
\end{equation}
which averages over all $10$ distinct output pairs per observation and subsequently across observations. Lower values indicate greater repeatability, with identical outputs yielding zero variation. As reported in Table~\ref{tab:rendering_precision}, InfiNoVA achieves lower pairwise perceptual variation than the compared stochastic synthesis baselines, indicating greater consistency across repeated generations under unchanged conditioning. We also report mean LPIPS against the corresponding real target-view observations, since a constant but incorrect output can achieve perfect repeatability. Together, these metrics assess both consistency and fidelity, supporting the reliability of our augmentations.
\vspace{-0.25cm}
\begin{table}[h]
\centering
\caption{Values report average pairwise LPIPS and mean LPIPS against ground truth over 5 generations for each of 5 observations.}
\label{tab:rendering_precision}
\footnotesize
\begin{tabular*}{\columnwidth}{@{\extracolsep{\fill}}lcc@{}}
\toprule
Method & Pairwise LPIPS $\downarrow$ & LPIPS to GT $\downarrow$ \\
\midrule
GLD~\cite{jang2026gld} & 0.5428 & 0.4810 \\
VISTA~\cite{tian2025view} & 0.2765 & 0.5002 \\
\textbf{InfiNoVA (Ours)} & \textbf{0.1406} & \textbf{0.466} \\
\bottomrule
\end{tabular*}
\vspace{-0.25cm}
\end{table}

\subsection{Policy Evaluation}
\begin{figure}[h]
    \centering
    \includegraphics[width=\linewidth]{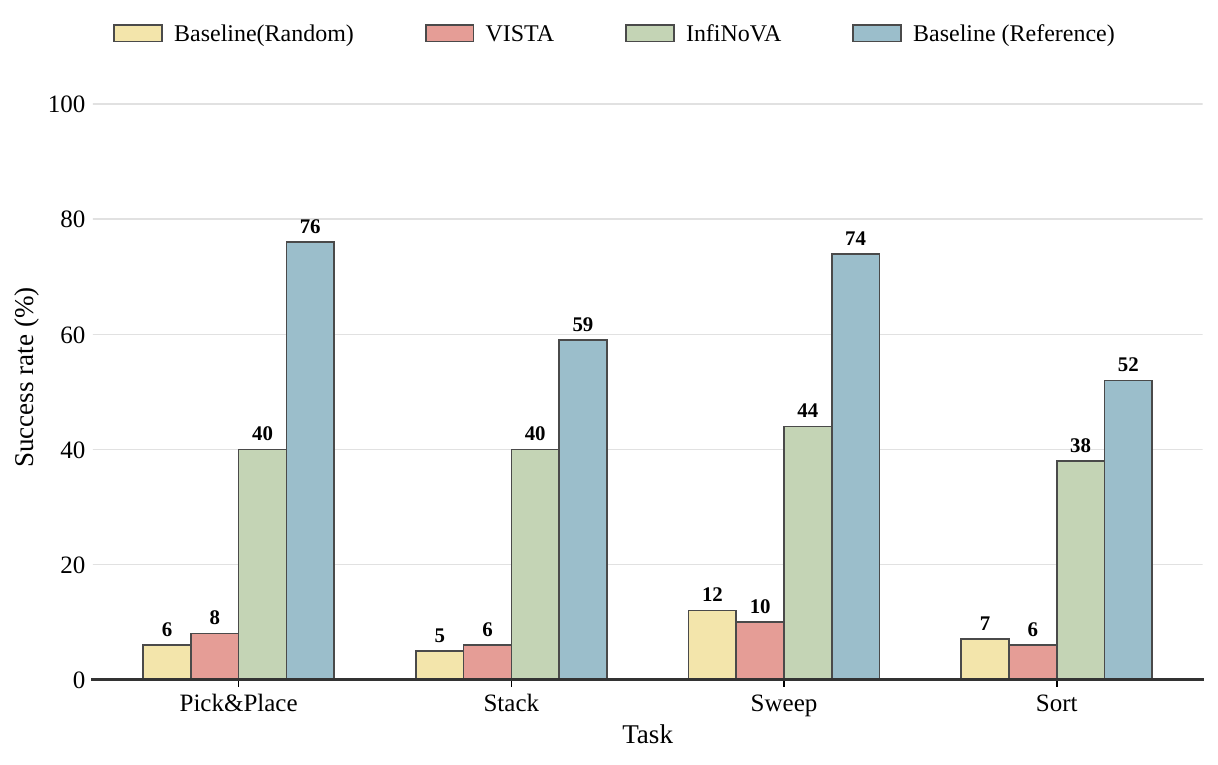}
    \caption{Policy success rates across four manipulation tasks.}
    \label{fig:policy_eval}
    \vspace{-0.3cm}
\end{figure}

We evaluate whether novel-view augmentation improves policy robustness to camera viewpoint shifts across four manipulation tasks: Pick-and-Place, Stack, Sweep, and Sort. Figure~\ref{fig:policy_eval} reports success rates over 100 rollouts per result. Baseline (Reference) is trained and evaluated on the original reference viewpoint, whereas Baseline (Random) evaluates the same policy from randomized viewpoints to measure degradation under viewpoint shift. VISTA and InfiNoVA are trained on datasets augmented with views synthesized by their respective frameworks and evaluated using the same randomized-view protocol as Baseline (Random).

Averaged across the four tasks in Fig.~\ref{fig:policy_eval}, the Baseline (Reference) achieves an average success rate of 65.3\%. When the same single-view policy is evaluated from randomized camera viewpoints, average success drops to 7.5\%, corresponding to an 88.5\% relative degradation and highlighting its strong dependence on the training viewpoint.

VISTA-based augmentation provides no overall recovery, with an average success rate of 7.5\%. In contrast, InfiNoVA achieves 40.5\% average success, corresponding to a \(5.4\times\) improvement over VISTA under viewpoint perturbations. Relative to the single-view policy under camera shift, InfiNoVA recovers approximately 57\% of the performance lost with respect to the in-distribution reference.

The remaining gap to the Baseline (Reference) reflects the harder evaluation setting: the reference policy is trained and tested from the same fixed viewpoint, whereas InfiNoVA is evaluated across randomized camera poses. These results demonstrate substantially improved generalization to unseen viewpoints while retaining the original policy architecture.

\subsection{Beyond Multi-Camera Training}
\label{subsec:multicam}

To evaluate the benefit of viewpoint synthesis from simply adding more physical cameras, we compare three policies. The single-view baseline is trained using demonstrations from only the reference camera view. The multi-view baseline is trained on the same demonstrations recorded from all five physical camera views, without NVS augmentation. Our method augments these demonstrations with densely synthesized novel views. All policies are evaluated over 100 Pick-and-Place trials using the same unseen camera poses with randomized viewpoints and distances. The single-view baseline achieves 76\% success on its training view but drops to 6\% on random views. Multi-view training improves performance to 24\%, while InfiNoVA increases success to 40\%. These results show that dense viewpoint augmentation provides stronger generalization than training on a sparse set of physical views.

\vspace{-0.2cm}
\section{conclusion}
We presented InfiNoVA, a novel-view augmentation framework that converts sparse synchronized multi-camera demonstrations into a dense distribution of geometrically grounded training views. By reconstructing manipulation trajectories as time-varying Gaussian Splat representations, InfiNoVA preserves task-relevant scene structure and state--action correspondence while varying the camera viewpoint. Across four real-world manipulation tasks, InfiNoVA produces higher-fidelity, more temporally consistent, and more repeatable novel views than existing synthesis approaches. These improvements translate directly to downstream robustness as under unseen randomized viewpoints, InfiNoVA-trained policies achieve $5.4\times$ higher average success than both VISTA-based augmentation and the unaugmented policy. This shows that viewpoint diversity alone is insufficient; synthetic views must also preserve the underlying task state to avoid contradictory state--action supervision. InfiNoVA further achieves $1.7\times$ higher success than training directly on all five physical camera views, demonstrating the benefit of dense synthesized viewpoint coverage beyond sparse multi-camera training.

The current study focuses on a controlled real-world setting with synchronized multi-camera observations, where the available viewpoints provide strong geometric coverage of the manipulation workspace. While our experiments demonstrate robust generalization across randomized unseen viewpoints, extending the sampling range further beyond the observed workspace would be an interesting direction for future study. Our evaluation considers four manipulation tasks on a single robot platform with SmolVLA, providing a consistent setting for isolating the effect of viewpoint augmentation. Evaluating the approach across additional embodiments, policy architectures, and more diverse long-horizon tasks would further establish its
generality. Finally, the current pipeline reconstructs scene representations offline at each timestep; future work could explore temporally shared reconstruction or more compact scene representations to further improve scalability.

\bibliographystyle{IEEEtran}
\bibliography{sections/reference}

\end{document}